\documentclass{article} 
\usepackage{nips14submit_e,times}

\nipsfinalcopy 

\usepackage[ruled,vlined]{algorithm2e}
\usepackage{algorithmic}
\usepackage{amsfonts}
\usepackage{amsmath}
\usepackage{amssymb}
\usepackage{bm}
\usepackage{color}
\usepackage{graphicx}
\usepackage{mathrsfs}
\usepackage{tikz}
\usepackage{url}

\usepackage[font=small,labelfont=bf,tableposition=top]{caption}
\usepackage[font=footnotesize]{subcaption}

\usetikzlibrary{arrows,calc,shapes,snakes,automata,backgrounds,fit,petri}

\def\figures{1}

\allowdisplaybreaks[1]

\def\argmin{\mathop{\rm argmin}}

\def\tr{\mathop{\sf tr}}
\def\trace{\mathop{\sf trace}}

\def\Ebb{\mathbb{E}}
\def\Ibb{\mathbb{I}}
\def\Rbb{\mathbb{R}}

\def\balpha{\bm{\alpha}}
\def\bbeta{\bm{\beta}}

\def\bmu{\bm{\mu}}

\def\bpi{\bm{\pi}}

\def\bpsi{\bm{\psi}}

\def\btheta{\bm{\theta}}
\def\bvartheta{\bm{\vartheta}}

\def\bxi{\bm{\xi}}
\def\bzeta{\bm{\zeta}}

\def\0{\mathbf{0}}
\def\1{\mathbf{1}}

\def\b{\mathbf{b}}
\def\c{\mathbf{c}}

\def\m{\mathbf{m}}

\def\s{\mathbf{s}}
\def\t{\mathbf{t}}
\def\u{\mathbf{u}}
\def\v{\mathbf{v}}

\def\y{\mathbf{y}}
\def\z{\mathbf{z}}

\def\I{\mathbf{I}}

\def\P{\mathbf{P}}

\def\U{\mathbf{U}}
\def\V{\mathbf{V}}

\def\Dcal{\mathcal{D}}
\def\Gcal{\mathcal{G}}

\def\Lcal{\mathcal{L}}

\def\Ncal{\mathcal{N}}
\def\Ocal{\mathcal{O}}

\def\erm{\mathrm{e}} 
\def\true{\mathsf{T}}
\def\false{\mathsf{F}}
\def\data{\mathscr{D}}
\def\diag{\mathsf{diag}}

\def\defined{\stackrel{.}{=}}
\newcommand{\appropto}{\mathrel{\vcenter{
  \offinterlineskip\halign{\hfil$##$\cr
    \propto\cr\noalign{\kern2pt}\sim\cr\noalign{\kern-2pt}}}}}

\definecolor{Blue}{rgb}{0.0,0.0,1.0}

\title{Scalable Bayesian Modelling of Paired Symbols}

\author{
Ulrich Paquet \\
Microsoft Research \\
Cambridge, United Kingdom 
\And
Noam Koenigstein \\
Microsoft R\&D \\
Herzliya, Israel 
\And
Ole Winther \\
Technical University of Denmark \\
Lyngby, Denmark 
}

\begin{document} 

\maketitle

\begin{abstract} 
We present a novel, scalable and Bayesian approach to modelling the occurrence of pairs of symbols $(i,j)$ drawn from a large vocabulary.
Observed pairs are assumed to be generated by a
simple popularity based selection process followed by censoring using a preference function.
By basing inference on the well-founded principle of variational bounding, and using new site-independent bounds, we show how a scalable inference procedure can be obtained for large data sets.
State of the art results are presented on real-world movie viewing data.
\end{abstract} 

\section{Introduction}
\label{sec:introduction}

We wish to model the occurrence of pairs of discrete symbols $(i,j)$ from a finite set, or predict the occurrence of symbol $j$
given that the other symbol is $i$. These pairs might be tuples of $(\textrm{\emph{user}}, \textrm{\emph{item}})$
purchase events, or a stream of $(\textrm{\emph{user}}, \textrm{\emph{game}})$ gameplay events.
From such a model, a recommender system can be tailored around the conditional probability of item or game $j$,
given user $i$.
Alternatively, these tuples might be $(\textrm{\emph{word}}_1, \textrm{\emph{word}}_2)$ bigrams in a simple language model.
If there are $I$ and $J$ of each symbol, their discrete density can be fully modelled 
as a multinomial distribution with $I \times J$ normalized counts, one for each pair.
In practice, data is typically sparse compared to large symbol vocabulary sizes, with $I \approx 10^7$ and $J \approx 10^5$ in tasks considered in this paper, and this prevents the full multinomial from generalizing:
from user $i$ watching only one movie $j$, we would like to infer the odds of her viewing other movies $j'$.
This necessitates more compactly parameterized models, which commonly associate real-valued vectors $\u_i \in \Rbb^{K}$ and $\v_j \in \Rbb^{K}$
(where $K \ll I,J$) with user $i$ and item $j$, and draws on an energy $\u_i^T \v_j$ to couple them.

This paper proposes a new approach to modelling the occurrence of pairs of symbols, 
and makes two main contributions.
First, pairwise data is modelled through a simple selection process followed by a preference function that censors the data: in the generative process, pairs $(i,j)$ are presented to a censor at a basic rate, which then chooses to include them in the stream of data with odds that depend only on the coupling energy $\u_i^T \v_j$.
Inference is based on the well-founded principle of variational bounding.
Second, we show how a scalable procedure can be obtained by using novel looser site-independent bounds.

To see why scalability is a challenge, consider 
the bilinear softmax distribution  
\begin{equation} \label{eq:softmax}
p(i,j) = \erm^{\u_i^T \v_j} \Big/ \sum_{i', j'} \erm^{\u_{i'}^T \v_{j'}} \ ,
\end{equation}
whose normalizing constant sums over all $I \times J$ discrete options.
When $i$ is given, $p(j|i)$ defines softmax regression, the multi-class extension of logistic regression.
The bilinear softmax function poses a practical difficulty:
the large sums from the normalizing constant appear in the likelihood gradient through
%
$\partial \log p(i,j) / \partial \u_i = \v_j - \sum_{j'=1}^{J} w_{ij'} \v_{j'}$,
where
$w_{ij} \defined \erm^{\u_i^T \v_j} / \sum_{i', j'} \erm^{\u_{i'}^T \v_{j'}}$
requires a sum over all $IJ$ pairs in its normalizer.
On observing a pair $(i,j)$, the likelihood is increased by
pulling $\u_i$ towards $\v_j$, while
simultaneously pushing it further from all other $\v_{j'}$.
There were recent approaches to using the softmax function at scale. Mnih and Teh \cite{Mnih_2012} used noise contrastive estimation \cite{JMLR:gutmann12a} to estimate the expensive softmax gradients when training neural probabilistic language models, which improves on using
importance sampling for gradient estimation \cite{Bengio03quicktraining}.
In a different approach, the normalization problem can be addressed by redefining $p(j|i)$ as a tree-based hierarchy of smaller softmax functions; this has a direct application to implicit-feedback collaborative filtering \cite{Mnih_2012_label}.
Alternatively, modelling can be done by formulating a simpler objective function based on a classification likelihood,
and including stochastically ``negative sampled'' pairs during optimization.
This was done for skip-gram models that consider $(\textrm{\emph{word}}_1, \textrm{\emph{word}}_2)$ pairs \cite{Mikolov2013},
and for $(\textrm{\emph{user}}, \textrm{\emph{item}})$ pairs \cite{Paquet_2013},
where the latter work assumed that each pair can appear at most once.
There additionally exists a large body of tangential work, which
models an i.i.d.~observation \emph{given} $i$ and $j$, instead of doing density estimation as described above.
These include the stochastic block model and its extensions for binary matrices or graphs \cite{Airoldi_2008} and the family of ``probabilistic matrix factorization'' models for a variety of likelihood functions for the observation \cite{DBLP:journals/corr/GopalanHB13, Marlin2009, MnihPMF}. 
The restriction of each pair to appearing at most once places us in the domain of one-class matrix completion \cite{Hu_2008,Pan_2009,Sindhwani_2010},
where modelling is typically done by formulating different loss functions over the absent pairs (or missing values in the matrix).
In these set-ups, a cost value is typically associated with each possible pair. These can be predefined \cite{Hu_2008,Pan_2009} or optimized for \cite{Sindhwani_2010}.

This paper has large-scale collaborative filtering and recommender systems in mind, and places two requirements on the model and inference procedure that do not coexist in other work. {\bf 1.}~Crucially, \emph{inference should scale with $D$}, the size of the dataset, i.e.~the number of observed pairs, and not with $IJ$, the number of possible pairs.
{\bf 2.}~We prefer a Bayesian approach that incorporates parameter uncertainty in our inference.
This is particularly useful when data is scarce; if game $j$ was played by a handful of users, its lack of usage should be reflected in the posterior estimate of parameters associated with $j$.
To fulfil these requirements,
we borrow an unorthodox idea from 
\cite{Paquet_2013, DBLP:journals/corr/PaquetK13}, which views the stream of data as a censored one.
Their perspective is that of a one-class model, which contrasts the observations against an unobserved \emph{``negative background''}, although unlike \cite{Paquet_2013}, a pair $(i,j)$ can 
repeatedly be observed.
In Section \ref{sec:scalable}, this ``negative background'' will be employed in various caches as part of the inference pipeline.
Our approach practically improves on that of \cite{Paquet_2013}, where the data set size was effectively doubled
as the non-revealed stream was stochastically resampled and averaged over.
As the ``non-revealed half'' continually changed due to resampling, the inference procedure also did 
not comfortably map to a distributed architecture.
Because exact inference in our model is not possible, we resort to approximating the parameter posterior via a variational lower bound.
In this Variational Bayes setting, with a fully factorized posterior approximation, the bound is iteratively maximized through closed-form updates of each factor. The updates are in terms of natural gradients, and are embarrassingly parallel.
Empirically, our approach achieves state of the art results on two large-scale recommendation problems (Section \ref{sec:eval}).

\section{Generative model for pairs with censoring}

A pair $(i,j)$ will be represented as a pair $(\y, \z)$ of binary indicator vectors, where only bits $i$ and $j$ are ``on'' in $\y \in \{0, 1 \}^I$ and $\z \in \{ 0, 1 \}^J$ respectively. 
We shall model the data stream by appending a binary variable $o = \true$ (true) to each pair: we \emph{did} observe that symbols $i$ and $j$ co-occurred, user $i$ played game $j$ today, and so on.
We therefore observe a stream of $D$ pairs, which takes the form $\{ o_d = \true, \y_d, \z_d \}_{d= 1}^{D}$.
The censored approach assumes that there were a number of pairs that did not surface in the data stream, such that $o = \false$ (false). \emph{We do not know which pairs and how many they were}, but in practice we will allow the length of the censored stream be specified as a hyperparameter $D'$,
and assume that
$\{ o_{d'} = \false \}_{d' = 1}^{D'}$
is additionally provided.
Let data $\data \defined \{ \{ o_d = \true, \y_d, \z_d \}_{d=1}^{D}, \{ o_{d'} = \false \}_{d'=1}^{D'} \}$ denote all observations.
The ratio $D / D'$ can be seen as a pre-specified positive to negative class ratio;
various settings of $r$ in $D' = rD$ are investigated in Section \ref{sec:eval}.
The censored stream constitutes the
``negative background'' against which the energy $\u_i^T \v_j$ will be fit, and
it plays a role similar to that of the softmax normalizer in
the gradient of $\log p(i,j)$ from (\ref{eq:softmax}):
on observing a pair $(i,j)$, $\u_i$ is pulled towards $\v_j$ and pushed further from all other $\v_{j'}$.

We additionally associate real-valued biases $b_i$ (and $b_j$) with each user and item, modifying the energy to $\u_i^T \v_j + b_i + b_j$.
They play a useful interpretive role in distinguishing between polarizing and non-polarizing content
in a recommender system: content that appeals
to a wide range of tastes is described by a $\v_j$ with smaller norm, and their appeal is modelled by a positive taste-independent bias. Polarizing content is described by a large-normed $\v_j$ and a negative taste-independent bias; it is only enjoyed by a narrow sliver of tastes.

We propose a model which combines popularity-based selection with a personalized preference function to model $(i,j)$. \textbf{1.}~In a \emph{selection step} a user $i$ is chosen with probability $\pi_i$, and an item $j$ is chosen with probability $\psi_j$. \textbf{2.}~In a 
\emph{censoring step} the pair $(i,j)$ is observed with probability $\sigma( \u_i^T \v_j+ b_i + b_j )$ and censored with probability $1-\sigma( \u_i^T \v_j+ b_i + b_j )$,
where $\sigma(a) = 1 /(1+e^{-a})$ is the logistic function.

\begin{figure}[h]
\begin{center}
\ifnum\figures=1
\begin{tikzpicture}[bend angle=45,>=latex]
\tikzstyle{obs} = [ circle, thick, draw = black!100, fill = black!20, minimum size = 7mm ]
\tikzstyle{lat} = [ circle, thick, draw=black!100, fill = red!0, minimum size = 7mm ]
\tikzstyle{par} = [ circle, draw, fill = black!100, minimum width = 3pt, inner sep = 0pt]	
\tikzstyle{every label} = [black!100]

\begin{scope}[node distance = 1.5cm and 1.5cm,rounded corners=4pt]
\node [obs] (t) {$\true$}; 
\node [obs] (f) [ right of = t ] {$\false$};
\node [obs] (zo) [ above of = t]  {$\z_{d}$}
       edge [post] (t);
\node [obs] (yo) [ left of = zo]  {$\y_{d}$}
       edge [post] (t);
\node [lat] (yh) [ above of = f]  {$\y_{d'}$}
       edge [post] (f);
\node [lat] (zh) [ right of = yh]  {$\z_{d'}$}
       edge [post] (f);
\node [lat] (pi) [ above of = zo] {$\bpi$};
       \draw[-latex] (pi) to[out=220,in=90] (yo);
       \draw[-latex] (pi) to[out=320,in=90] (yh);
\node [lat] (psi) [ above of = yh] {$\bpsi$};
       \draw[-latex] (psi) to[out=220,in=90] (zo);
       \draw[-latex] (psi) to[out=320,in=90] (zh);
\node [lat] (bu) [ below of = t] {$b_i$}
       edge [post] (t)
       edge [post] (f);
\draw (-1.5,-1.5) .. controls (-1.5,-0.5) and (0,-1.3) .. (0,-0.6);
\draw (-1.5,-1.5) .. controls (-1.5,-0.5) and (0,-1.3) .. (0.9,-0.6);
\node [lat] (u) [ left of = bu ] {$\u_i$};
\node [lat] (v) [ below of = f ] {$\v_j$}
       edge [post] (f)
       edge [post] (t);
\draw (3.0,-1.5) .. controls (3.0,-0.5) and (1.5,-1.3) .. (1.5,-0.6);
\draw (3.0,-1.5) .. controls (3.0,-0.5) and (1.5,-1.3) .. (0.6,-0.6);
\node [lat] (bv) [ right of = v] {$b_j$};
\draw (-2.0,-0.3) node {$D$};
\draw (3.5,-0.3) node {$D'$};
\draw (-2.0,-2.0) node {$I$};
\draw (3.5,-2.0) node {$J$};
				
\begin{pgfonlayer}{background}
\filldraw [line width = 2pt, draw=black!30, fill=white!100]
(-2.3cm,2.3cm) rectangle (0.6cm,-0.6cm)
(0.9cm,2.3cm) rectangle (3.8cm,-0.6cm)
(-2.3cm,-0.9cm) rectangle (0.6cm,-2.3cm)
(0.9cm,-0.9cm) rectangle (3.8cm,-2.3cm);
\end{pgfonlayer}							
\end{scope}
\end{tikzpicture}	
\else
\emph{figure here}
\fi
\end{center}
\caption{A generative model for observing $D$ pairs of symbols, assuming that $D'$ \emph{unknown} pairs were censored.
}
\label{fig:graphicalmodel}
\end{figure}
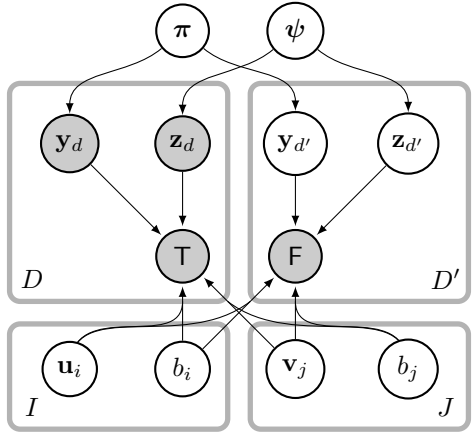
 
Let $\U \defined \{ \u_i \}_{i=1}^I$ and $\V \defined \{ \v_j \}_{j=1}^J$ denote all bilinear parameters
and $\b \defined \{ \{ b_i \}_{i=1}^I,  \{ b_j \}_{j=1}^J \}$ denote biases,
with $\bzeta \defined \{ \U, \V, \b \}$. Lastly $\bvartheta \defined \{ \bzeta, \bpi, \bpsi \}$ includes multinomial parameters $\bpi$ and $\bpsi$.
The generative process is illustrated in Figure \ref{fig:graphicalmodel}, and is as follows:
draw parameters $\bvartheta$ from their prior distributions (given explicitly below). Repeat drawing pairs $(i,j)$ with indexes drawn from ${\rm Discrete}(\bpi)$ and ${\rm Discrete}(\bpsi)$ and observe the pairs with probability $\sigma( \u_i^T \v_j+ b_i + b_j )$.
$D$ such pairs are seen, while we assume that $D'$, the number of censored data points, is specified as a hyperparameter.
The density of an uncensored data point $d$ is therefore
\[
p(o_d = \true, \y_d, \z_d | \bvartheta) = p(o_d = \true | \y_d, \z_d, \bzeta) \, p(\y_d | \bpi) \, p(\z_d | \bpsi) 
= \prod_{i,j} \big[ \pi_i \, \psi_j \, \sigma( \u_i^T \v_j+ b_i + b_j ) \big]^{y_{di} z_{dj}} ,
\]
while $p(o_{d'} = \false | \y_{d'}, \z_{d'}, \bzeta) 
= \prod_{i,j} (1 - \sigma( \u_i^T \v_j+ b_i + b_j ) )^{y_{d'i} z_{d'j}}$
is the odds of censoring pair $d'$ if its indexes were known.
The censored indexes $\y_{d'}$ and $\z_{d'}$ are unknown; by including their prior and marginalizing over them,  $p(o_{d'} = \false | \bvartheta)$ is a mixture of $I J$ components.

The joint density of $\data$ and the unobserved variables $\btheta \defined \{ \bvartheta, \{ \y_{d'}, \z_{d'} \}_{d'=1}^{D'} \}$ depends on further priors on $\bvartheta$,
for which we choose Dirichlet priors for
$p(\bpi) = \Dcal(\bpi ; \balpha_0)$ and 
$p(\bpsi) = \Dcal(\bpsi ; \balpha_0)$.
The other priors are fully factorized Gaussians, with
$p(\U) = \prod_i \Ncal(\u_{i} ; \0, \tau_u^{-1} \I)$ and
$p(\V) = \prod_j \Ncal(\v_{j} ; \0, \tau_v^{-1} \I)$ and,
with some overloaded notation, $p(\b) = \prod_i \Ncal(b_{i} ; 0, \tau_b^{-1}) \prod_j \Ncal(b_{j} ; 0, \tau_b^{-1})$.
The hierarchical model could be extended further with Gamma hyperpriors on the various Gaussian precisions $\tau$,
or Normal-Wishart hyperpriors on both of the Gaussian parameters \cite{Paquet_2012, Salakhutdinov_2008}.
If the symbols $i$ and $j$ were accompanied by meta-data tags, these could also be incorporated into the Bayesian model \cite{Koenigstein2013}. For the sake of clarity, we omit these additions in this paper.
The joint density decomposes as
\begin{align}
p(\data , \btheta) & =
p(\data | \{ \y_{d'}, \z_{d'} \}, \bvartheta) \, p(\{ \y_{d'}, \z_{d'} \} | \bpi, \bpsi) \, p(\bvartheta) \nonumber \\
& =  \prod_{i,j} \sigma ( \u_i^T \v_j + b_i + b_j )^{c_{ij}} [ 1 - \sigma ( \u_i^T \v_j + b_i + b_j )]^{\sum_{d'} y_{d' i} z_{d' j}} \nonumber \\ 
& \quad \cdot \prod_i \pi_i^{c_i + \sum_{d'} y_{d' i}} \cdot \prod_j \psi_j^{c_j + \sum_{d'} z_{d' j}} \cdot p(\U) \, p(\V) \, p(\b) \, p(\bpi) \, p(\bpsi) \ , \label{eq:joint}
\end{align}
where the uncensored data likelihood was regrouped using observation counts
$c_{ij} \defined \sum_d y_{di} z_{dj} \in \{ 0, 1, 2, \ldots, D \}$ for each pair $(i,j)$,
and marginal counts $c_i \defined \sum_d y_{di} $ and $c_j \defined \sum_d z_{dj}$.
Note that $\sum_{i,j} c_{ij} = D$.
Marginalizing $p(\mathscr{D} , \btheta)$ over $\{ \y_{d'}, \z_{d'} \}$ gives a mixture of ${D' + IJ - 1}\choose{IJ - 1}$ components, each representing a different way of assigning $D'$ indistinguishable $\false$'s to $I J$ distinguishable bins, or assigning nonnegative counts $c_{ij}'$ with $\sum_{i,j} c_{ij}' = D'$ to a ``negative class count matrix''.

At first glance of (\ref{eq:joint}), it would seem as if inference would still scale with $I J$, and that we have done nothing more than
match the bilinear softmax's $\Ocal(I J)$ computational burden through the introduction of $D'$. The following sections are devoted to developing a variational approximation, and with it a practically scalable inference scheme that relies on various ``negative background'' caches.

\section{Variational Bayes} \label{sec:vb}

To find a scalable yet Bayesian inference procedure, we approximate $p(\btheta | \data)$ with a factorized surrogate density $q(\btheta)$, found by maximizing a variational lower bound to $\log p(\data)$ \cite{Waterhouse1995}.
First, we lower-bound each logistic function in (\ref{eq:joint}) by associating a parameter $\xi_{ij}$ with it \cite{Jaakkola_Jordan_1996}. Dropping subscripts, each bound would be
$
\sigma(\pm a) \ge \sigma(\xi) \exp ( - \lambda(\xi) \left(a^2 - \xi^2\right) \pm \frac{a}{2} - \frac{\xi}{2} )
$,
where the lower bound on $1 - \sigma(a)$ is that of $\sigma(-a)$ above.
The bound depends on the deterministic function $\lambda(\xi) \defined \frac{1}{2 \xi} [\sigma(\xi) - \frac{1}{2}]$.
Let $\bxi \defined \{ \xi_{ij} \}$ denote the set of logistic variational parameters,
and substitute the bound into (\ref{eq:joint}) to get $p(\data, \btheta) \ge p_{\bxi}(\data, \btheta)$.
Our variational objective $\Lcal_{\bxi}[q]$, as a function of $\bxi$ and functional of $q$, follows from
\begin{equation}
\log p(\data) \ge \log \int p_{\bxi}(\data, \btheta) \, \mathrm{d} \btheta
\ge \int q(\btheta) \log \frac{p_{\bxi} (\data, \btheta)}{q(\btheta)} \, \mathrm{d} \btheta \defined \Lcal_{\bxi}[q] \ ,
\label{eq:L}
\end{equation}
which will be maximized with respect to $q$ and $\bxi$.
The factorization of $q$ employed in this paper is
\begin{equation} 
q(\btheta) \defined \prod_{i} q(b_i) \prod_{k} q( u_{ik}) \cdot \prod_{j} q(b_j) \prod_{k} q(v_{jk})
\cdot \prod_{d'} q(\y_{d'}) \, q(\z_{d'}) \cdot q(\bpi) \, q(\bpsi) \ .
\label{eq:factorizing}
\end{equation}
The factors approximating each symbol's features in $\U$, $\V$, and $\b$ are chosen to be a Gaussian, for example $q(u_{ik}) = \Ncal ( u_{ik} ;  \mu_{ik}, \omega_{ik}^{-1} )$.
The approximating factors $q(\bpi)$ and $q(\bpsi)$ are both Dirichlet, for example $q(\bpi) = \Dcal(\bpi ; \balpha)$.
The bound in (\ref{eq:L}) is stated fully in Appendix \ref{sec:bounddetails}.

For the purpose of obtaining a scalable algorithm, the most important parameterizations are for the
categorical (discrete) factors $q(\y_{d'})$ and $q(\z_{d'})$.
We shall argue and show in Sections \ref{sec:scalable} 
and \ref{sec:eval} that choosing $D' \approx D$ is desired,
and as $D'$ is potentially large, the \emph{parameters of $q(\y_{d'})$ will be tied}.
\emph{This tying of parameters is the key to achieving a scalable algorithm.}
We let all $q(\y_{d'})$ share the same parameter vector $\s$ on the probability simplex, such that
$
q(\y_{d'}) = \prod_i s_i^{y_{d' i}}
$
for all $d'$. Similarly, all $q(\z_{d'})$ share probability vector $\t$, such that $q(\z_{d'}) = \prod_j t_j^{z_{d' j}}$ for all $d'$.

\paragraph{Making and trading predictions}

Our original desideratum was to infer the probability of symbol $j$, conditional on the other symbol being $i$, and the observed data.
Bayesian marginalization requires us to average the predictions over the model parameter posterior distribution.
Here it is an analytically intractable task,
which we approximate by using $q$ as a surrogate for the true posterior.
Firstly,
$
p(o = \true | \y, \z, \data) \approx \int p(o = \true | \y, \z,  \bvartheta)  q(\bvartheta) \, \mathrm{d} \bvartheta
= \int \sigma(a_{ij}) \, \Ncal (a_{ij} ; \mu_{ij}, \sigma_{ij}^2 ) \, \mathrm{d} a_{ij} \approx \sigma (x_{ij})
$ 
if $y_i = z_j = 1$.
The random variable $a_{ij}$ was defined as $a_{ij} \defined \u_i^T \v_j + b_i + b_j$, with its density approximated with its first two moments under $q$, i.e.~$\mu_{ij} \defined \Ebb_q[a_{ij}]$ and $\sigma_{ij}^2 \defined \Ebb_q[(a_{ij} - \mu_{ij})^2]$.
With $x_{ij} \defined \mu_{ij} \, / (1 + \pi \sigma_{ij}^2 / 8)^{1/2}$, the final approximation of a logistic Gaussian integral follows from \cite{MacKay_1992}.
Again using $q$, the posterior density of symbol $j$, provided that the first symbol is $i$, is approximately proportional to
(writing ``$\true$'' for ``$o = \true$'' for brevity)
\begin{equation}
p(z_j = 1 | \true, y_i=1, \data) \, \appropto \, p(\true | y_i=z_j=1, \data) \int p(z_j= 1| \bpsi) q(\bpsi) \, \mathrm{d} \bpsi
= \sigma (x_{ij}) \, \Ebb_q[ \psi_j ] \  . \label{eq:prediction}
\end{equation}
Hence $p(z_j = 1 | o = \true, y_i=1, \data) \approx \sigma (x_{ij}) \, \Ebb_q[ \psi_j ] \Big/ \sum_{j'} \Ebb_q[ \psi_{j'} ] \, \sigma (x_{ij'})$, normalizing to one.

\section{Scalable inference} \label{sec:scalable}

A scalable update procedure for the factors of $q(\btheta)$ is presented in this section, culminating in Algorithm \ref{alg}.
The algorithm optimizes over $t_{\max}$ loops, but can also be run until complete convergence as the evidence lower bound $\Lcal$
from (\ref{eq:L}) can be explicitly calculated.
We use \textbf{\textit{p}for} to indicate embarrassingly parallel loops, although the updates for $\s$, $\t$, and $\xi^*$ also make extensive use of parallelization.

Let graph $\Gcal = \{ (i,j) : c_{ij} > 0 \}$ be the sparse set of all observed pair indexes.
As there are $IJ$ logistic variational parameters $\xi_{ij}$, we shall divide them into two sets, those with indexes in $\Gcal$, and those without.
Therefore $\xi_{ij}$ shall be optimized for when $(i,j) \in \Gcal$,
while the $\xi_{ij}$'s shall \emph{share the same parameter value $\xi^*$} for $(i,j) \notin \Gcal$.
Even though the form of (\ref{eq:joint}) suggests that we would need two versions of $\xi_{ij}$, one for the bounded $\sigma$-term, and one its opposite, this is not required, as the optimization of the bound is symmetric. When $\xi_{ij}$ maximizes $\Lcal$ on the bounded $\sigma$-term, it simultaneously maximizes $\Lcal$ on the bounded $(1 - \sigma)$-term.
We'll use the shorthand $\lambda_{ij} \defined \lambda(\xi_{ij})$  for $(i,j) \in \Gcal$; similarly, $\lambda^*$ denotes $\lambda(\xi^*)$ when $(i,j) \notin \Gcal$.
The updates for symbols $i$ and $j$'s parameters mirror each other, and only the ``user updates'' are laid out in this section.

\begin{algorithm}[t]
\DontPrintSemicolon
\SetNlSty{}{}{:}
\SetAlgoNlRelativeSize{-2}
\SetKwFor{While}{\emph{p}for}{do}{end\emph{p}for} 
\textbf{input:} $\data$ (or $D'$ and all non-zero $c_{ij}$), $\balpha_0$, $\bbeta_0$, $\tau_u$, $\tau_v$, $\tau_b$ \;
\textbf{initialize:} $\xi^* \leftarrow 1$, $\s \leftarrow [1/I]$, $\t \leftarrow [1/J]$ \;
\For{$t=1 : t_{\max}$}{
update $q(\bpi)$ ;
update $q(\bpsi)$ ;
cache item-background $\P_{\ominus}$, $\m_{\ominus}^{\dagger}$, $\m_{\ominus}^{\ddagger}$, $\nu_{\ominus}$, $\varkappa_{\ominus}$ ; update $\s$ \;
\textbf{\emph{p}for} $i=1 : I$ \textbf{do} \{ update $q(b_i)$ ; update $\prod_{k=1}^{K} q(u_{ik})$ \} \;
cache user-background $\P_{\oplus}$, $\m_{\oplus}^{\dagger}$, $\m_{\oplus}^{\ddagger}$, $\nu_{\oplus}$, $\varkappa_{\oplus}$ ;
update $\xi^*$ ; update $\t$ \;
\textbf{\emph{p}for} $j=1 : J$ \textbf{do} \{ update $q(b_j)$ ; update $\prod_{k=1}^{K} q(v_{jk})$ \} \;
}
\caption{Paired Symbol Modelling\label{alg}}
\end{algorithm}

\paragraph{Gaussian updates for $q(u_{ik})$}

We will present here a bulk update of $\prod_k q(u_{ik})$, which is faster than sequentially maximizing $\Lcal_{\bxi}[q]$ with respect to each of them in turn.
We first solve for the maximum of $\Lcal$ with respect to a full Gaussian (not factorized)
approximation $\tilde{q}(\u_i) = \Ncal(\u_i ; \bmu_i, \P_i^{-1})$. The fully factorized $q(u_{ik})$ can then be recovered from the intermediate approximation
$\tilde{q}(\u_i)$ as those that minimize the Kullback-Leibler divergence $\mathsf{D}_{\mathrm{KL}}(\prod_k q(u_{ik}) \| \tilde{q}(\u_i))$: this is achieved when the
means of $q(u_{ik})$ match that of $\tilde{q}(\u_i)$, while their precisions match the diagonal precision of $\tilde{q}(\u_i)$.
The validity of the intermediate bound in proved in Appendix \ref{sec:traitupdates}.
The updates rely on careful caching, which we'll first illustrate through $\tilde{q}$'s precision matrix. $\Lcal$ is maximized when $\tilde{q}(\u_i)$ has as natural parameters a precision matrix
\begin{equation} \label{eq:Pi}
\P_i = 
\sum_{j \in \Gcal(i)} c_{ij} \cdot 2 \lambda_{ij} \cdot \Ebb_{q} \big[ \v_j \v_j^T \big]
+ \,
\overbrace{\sum_{d'} \sum_{j} \Ebb_q[y_{d' i} \, z_{d' j}] \cdot  2 \lambda_{ij} \cdot \Ebb_{q} \big[ \v_j \v_j^T \big]}^{\mathrm{(a)}} \, + \, \tau_u \I
\end{equation}
and a mean-times-precision vector $\m_i$, which we will state later.
Looking at $\P_i$ in (\ref{eq:Pi}), an undesirable sum over all $d'$ and $j$ is required in $\mathrm{(a)}$.
We endeavoured that the update would be \emph{sparse}, and only sum over observed indexes in $\Gcal(i) \defined \{ j : (i,j) \in \Gcal \}$.
The benefit of the shared variational parameters now becomes apparent. With $\Ebb_q[y_{d' i} \, z_{d' j}] = s_i t_j$ and
$\lambda_{ij} = \lambda^*$ when $(i,j) \notin \Gcal$, the sum in $\mathrm{(a)}$ decomposes as
%
\[
\mathrm{(a)} = 
\sum_{j \in \Gcal(i)} s_i t_j D' \cdot 2 (\lambda_{ij} - \lambda^*) \, \Ebb_{q} \big[ \v_j \v_j^T \big]
+ s_i D' \cdot 2 \lambda^{*} \cdot \overbrace{\sum_j t_j \Ebb_{q} \big[ \v_j \v_j^T \big]}^{\textrm{negative background $\P_{\ominus}$}}
 \ .
\]
Barring the ``negative background'' term, only a sparse sum that involves observed pairs is required.
This background term is rolled up into a global \emph{item}-background cache, which is computed once before updating all $q(u_{ik})$.
Throughout the paper, the $\ominus$ symbol will denote an item-background cache.
The cache
$ 
\P_{\ominus} \defined \sum_{j} t_j \, \Ebb_{q} [ \v_j \v_j^T ]
$ 
is used in each precision matrix update, for example
\[
\P_i = s_i D' \cdot 2 \lambda^{*} \cdot \P_{\ominus} + \sum_{j \in \Gcal(i)} \Big( c_{ij} \cdot 2 \lambda_{ij}
+ s_i t_j D' \cdot 2 (\lambda_{ij} - \lambda^{*}) \Big) \Ebb_{q} \big[ \v_j \v_j^T \big] + \tau_u \I \ . 
\]
We've deliberately laboured the above decomposition of an expensive update into a background cache and a sparse sum over actual observations, as it serves as a template for other parameter updates to come. Turning to the mean-times-precision vector $\m_i \defined \P_i \bmu_i$ of $\tilde{q}(\u_i)$, we find that
\begin{equation}
\m_i = \Ebb_{q}\left[ \sum_{j \in \Gcal(i)} c_{ij} \left(\tfrac{1}{2} - 2 \lambda_{ij} (b_i + b_j)\right) \v_j
+ \sum_{d'} \sum_{j} y_{d' i} \, z_{d' j} \left(- \tfrac{1}{2} - 2 \lambda_{ij} ( b_i + b_j ) \right) \v_j \right] . \label{eq:mp}
\end{equation} 
There is a subtle link between (\ref{eq:mp}) and the gradients of the bilinear soft-max likelihood, which we'll explore in the next paragraph.
To find $\m_i$, two additional caches are added to the item-background cache, and are computed once before any $q(u_{ik})$ updates. They are
$\m_{\ominus}^{\dagger} \defined \sum_{j} t_j \Ebb_{q} [ b_j ] \Ebb_{q} [ \v_j ]$
and $\m_{\ominus}^{\ddagger} \defined \sum_{j} t_j \Ebb_{q} [ \v_j ]$. 
The final mean-times-precision update is
\begin{align}
& \m_i = s_i D' \left[ \left(-\tfrac{1}{2} - 2 \lambda^{*} \Ebb_{q} \big[ b_i \big] \right) \m_{\ominus}^{\ddagger}  - 2 \lambda^{*} \m_{\ominus}^{\dagger} \right] \nonumber \\
& \qquad + \sum_{j \in \Gcal(i)} \Big( c_{ij} \left( \tfrac{1}{2} - 2 \lambda_{ij} \, \Ebb_{q} \big[ b_i  + b_j \big] \right) 
- s_i t_j D' \cdot 2 (\lambda_{ij} - \lambda^* ) \, \Ebb_{q} \big[ b_i+ b_j \big] \Big) \Ebb_{q} \big[ \v_j \big] \ , \label{eq:mtp-update}
\end{align} 
and again only sums over $j \in \Gcal(i)$ and not all $J$.
There are of course additional variational parameters $\xi_{ij}$, and they are computed and discarded when needed according to (\ref{eq:locallogistic}).

\paragraph{Bilinear softmax gradients}

The connection between this model and a bilinear softmax model can be seen when the biases are ignored.
Consider the gradient of $\Lcal$ with respect to \emph{mean} parameter $\bmu_i$,
\begin{equation} \label{eq:this-vs-softmax}
\nabla \Lcal(\bmu_i) = - \P_i \bmu_i + \frac{1}{2} \Bigg( \sum_{j \in \Gcal(i)} c_{ij} \Ebb_q\big[ \v_j \big] - D' \sum_{j} s_i t_j \Ebb_q \big[\v_j \big] \Bigg) \ .
\end{equation}
The gradient $\nabla \Lcal(\bmu_i)$ is zero at (\ref{eq:mp}), which was stated,
together with (\ref{eq:Pi}), in terms of \emph{natural} parameters.
As $\Lcal(\bmu_i)$ is quadratic, it can be exactly maximized; furthermore, the maximum with respect to $\P_i$ is
attained at the negative Hessian $\P_i = - \nabla^2 \Lcal(\bmu_i)$, given in (\ref{eq:Pi}).
The curvature of the bound, as a function of $\bmu_i$, directly translates into our posterior approximation's uncertainty of $\u_i$.
The log likelihood of a softmax model would be $L = \sum_d \log p(i_d, j_d)$, with the likelihood of each pair defined by (\ref{eq:softmax}).
The gradient of the log likelihood is therefore
\begin{equation} \label{eq:softmaxgradfull}
\nabla L(\u_i) = \sum_{j \in \Gcal(i)} c_{ij} \v_j - D \sum_{j} w_{ij} \v_j \ ,
\end{equation}
with weights $w_{ij} \defined \erm^{\u_i^T \v_j} / \sum_{i',j'} \erm^{\u_{i'}^T \v_{j'}}$ that sum to one over all $I J$ options.
The weights in (\ref{eq:this-vs-softmax}) were simply $s_i t_j$, and also sum to one over all options.
The difference between (\ref{eq:this-vs-softmax}) and (\ref{eq:softmaxgradfull}) is that $s_i t_j$ is used as a \emph{factorized substitute} for $w_{ij}$.
This simplification allows the convenience that none of the updates described in Section \ref{sec:scalable} need to be stochastic, and substitute functions, as employed by noise contrastive divergence to maximize $L$, are not required.
(The Hessian $\nabla^2 L(\u_i)$ contains a \emph{double-sum} over indexes $j$.)
Considering the two equations above, one might expect to set hyperparameter $D'$ to $D' = D$,
and in Section \ref{sec:eval} we show that this is a reasonable choice.

\paragraph{Gaussian updates for $q(b_i)$}

The maximum of $\Lcal$ with respect to $q(b_i)$ \emph{re-uses} cache $\m_{\ominus}^{\ddagger}$, but requires the additional cache
$
\nu_{\ominus} \defined \sum_{j} t_j \Ebb_{q} \big[ b_j \big]
$ 
to be precomputed. Gaussian $q(b_i)$ has a mean-times-pre\-ci\-sion parameter 
$\nu_i = s_i D' ( - \frac{1}{2} - 2 \lambda^* ( \nu_{\ominus} + \Ebb_{q} [ \u_i^T ]  \m_{\ominus}^{\ddagger} ) ) 
+ \sum_{j \in \Gcal(i)} \big( c_{ij} (\frac{1}{2} - 2 \lambda_{ij} \, \Ebb_{q} [ \u_i^T \v_j  + b_j ] ) 
-  s_i t_j D' \cdot 2 (\lambda_{ij} - \lambda^*) \, \Ebb_{q} [ \u_i^T \v_j +  b_j] \big)
$, 
%
and its precision parameter 
$\rho_i = 2 \lambda^* s_i D' + \sum_{j \in \Gcal(i)} \big( c_{ij} 2 \lambda_{ij} + s_i t_j D' 2 (\lambda_{ij} - \lambda^*) \big) + \tau_b$ 
follows a similar form.

\paragraph{Logistic bound parameter updates}

As discussed above, the logistic bound parameters $\xi_{ij}$ associated with observations $(i,j) \in \Gcal$ are treated individually whilst the remainder are shared and denoted by $\xi^*$. The individually optimized bound
\begin{equation} \label{eq:locallogistic}
\xi_{ij}^2 = \Ebb_q[ (\u_i^T \v_j + b_i + b_j)^2] 
\end{equation}
can be used anytime during the updates and then discarded (we always use the positive root for $\xi_{ij}$).
The shared parameter can be written in terms of cached quantities and a sum that scales with $D$ (the user-background cache is denoted with a $\oplus$ symbol, and mirrors the item-background cache):
\[
(\xi^*)^2 = \frac{1}{\mathcal{Z}} \Bigg( \tr \P_{\oplus} \P_{\ominus} + 2 \m_{\oplus}^{\ddagger T} \m_{\ominus}^{\dagger}
+ 2 \m_{\oplus}^{\dagger T} \m_{\ominus}^{\ddagger}
+ 2 \nu_{\oplus} \nu_{\ominus} + \varkappa_{\oplus} + \varkappa_{\ominus}
- \sum_{(i,j) \in \Gcal} s_i t_j \xi_{ij}^2 \Bigg)
\]
where $\mathcal{Z} \defined 1 - \sum_{(i,j) \in \Gcal} s_i t_j$.
Cache $\varkappa_{\ominus} \defined \sum_j t_j \Ebb_{q}[ b_j^2 ]$ also plays a role in the categorical updates.

\paragraph{Dirichlet updates}

As the multinomial distribution is conjugate to a Dirichlet, its updates have a particularly simple form. $q(\bpi)$ is Dirichlet $\Dcal(\bpi ; \balpha)$ with parameters
$
\alpha_i = \alpha_{0i} + c_i + s_i D'
$.
Each pseudo-count adds $c_i$, the number of views for user $i$, to the expected number of views that were censored and not made.

\paragraph{Categorical updates}

There are $D'$ categorical (discrete) factors $q(\y_{d'})$, and the key to finding a scalable inference procedure lies in tying all
their parameters together in $\s$, with $\sum_i s_i = 1$.
Looking at the second line of (\ref{eq:joint}), the factors depend on the expected bounded logistic functions
\begin{align*}
\Omega_{ij} & \defined \log \sigma(\xi_{ij}) - \lambda(\xi_{ij}) \Big( \Ebb_q[ (\u_i^T \v_j + b_i + b_j)^2] - \xi_{ij}^2 \Big)
 - \frac{\xi_{ij}}{2} - \frac{1}{2} \Ebb_q[\u_i^T \v_j + b_i + b_j] \ .
\end{align*}
The categorical parameters are, if we solve for all the $D'$ tied distributions $q(\y_{d'})$ \emph{jointly},
\[
\textstyle{ s_i \propto \exp \big( \Ebb_q [\log \pi_i] + \sum_j t_j \Omega_{ij} \big) }  \ .
\]
In practice, each entry $\log s_i + \mathrm{const}$ can be computed in parallel; afterwards, they are renormalized to give $\s$.
To find $\s$, an efficient way is needed to determine $\sum_j t_j \Omega_{ij}$, and this can again be done with careful bookkeeping.
The observed terms $j\in \Gcal(i)$ are treated differently from the rest. For observed terms we can use the 
optimal logistic parameters in (\ref{eq:locallogistic}) to simplify $
\Omega_{ij} \defined \log \sigma(\xi_{ij}) - \frac{\xi_{ij}}{2} - \frac{1}{2} \Ebb_q[\u_i^T \v_j + b_i + b_j]
$.
By denoting $\Omega_{ij}(\xi^*)$ evaluated with the shared parameter $\xi^*$ by $\Omega^*_{ij}$,
we can write
$
\sum_j t_j \Omega_{ij} =  \sum_{j\in \Gcal(i)} t_j (\Omega_{ij}-\Omega_{ij}^*) + \sum_j t_j \Omega_{ij}^*
$. 
The first term scales with $D$ and the second term can be written using cached quantities:
$ 
\sum_j t_j \Omega_{ij}^*  = - \lambda^* ( \tr \Ebb_q[ \u_i \u_i ^T ] \P_{\ominus} + 2 \Ebb_q[ b_i \u_i^T ] \m_{\ominus}^{\ddagger} 
 + 2 \Ebb_q[ \u_i^T ] \m_{\ominus}^{\dagger} + \Ebb_q[b_i^2] + 2 \Ebb_q[b_i] \nu_{\ominus} + \varkappa_{\ominus} ) 
+ \log \sigma(\xi^*)
 +\frac{ (\xi^*)^2}{2} \lambda^*   - \frac{\xi^*}{2} - \frac{1}{2} ( \Ebb_q[\u_i^T] \m_{\ominus}^{\ddagger} + \Ebb_q[b_i] + \nu_{\ominus} )
$.

\section{Evaluation} \label{sec:eval}

\begin{figure} [t]
\begin{center}
\ifnum\figures=1
\includegraphics[width=0.48\textwidth]{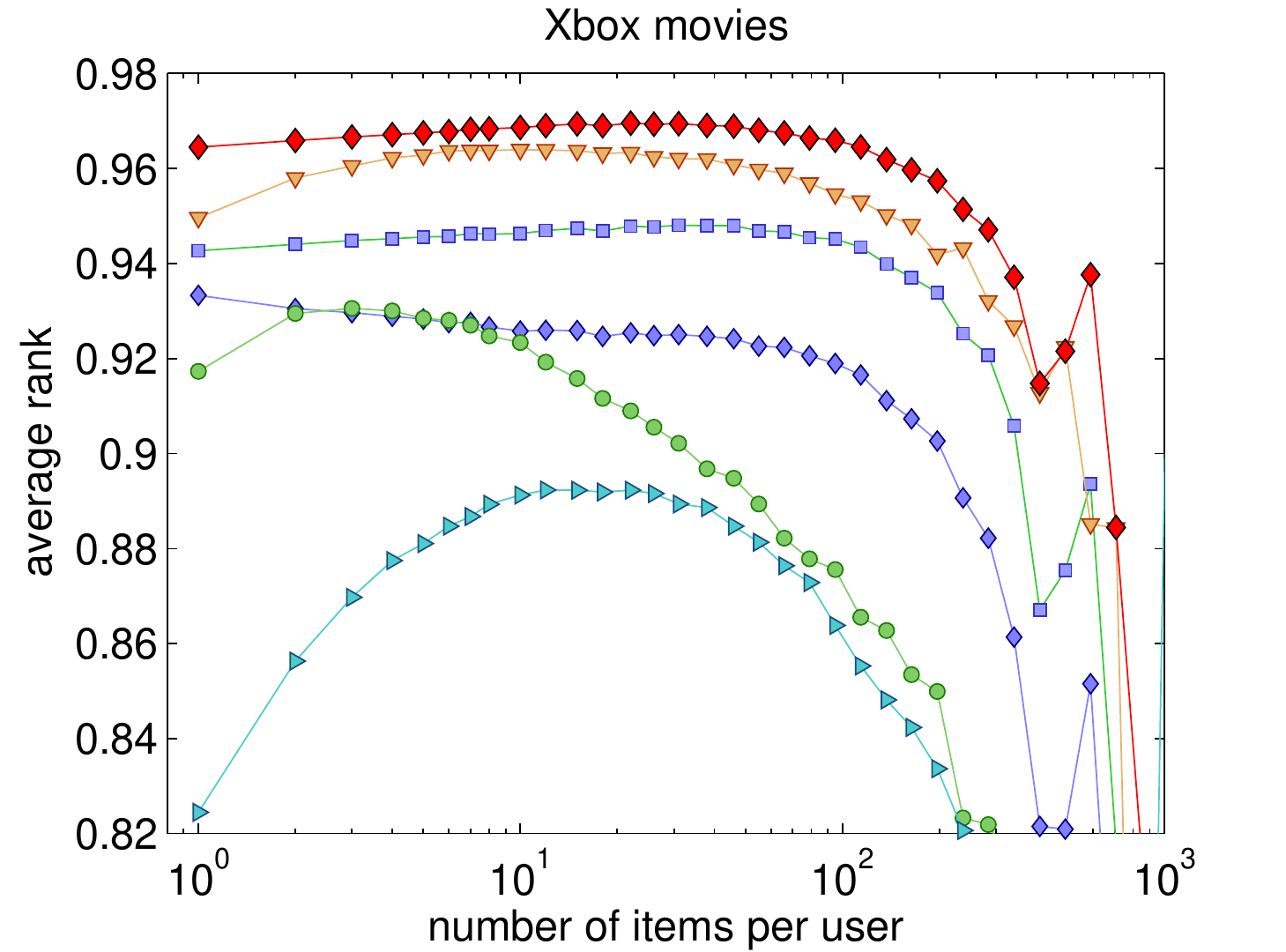}
\includegraphics[width=0.48\textwidth]{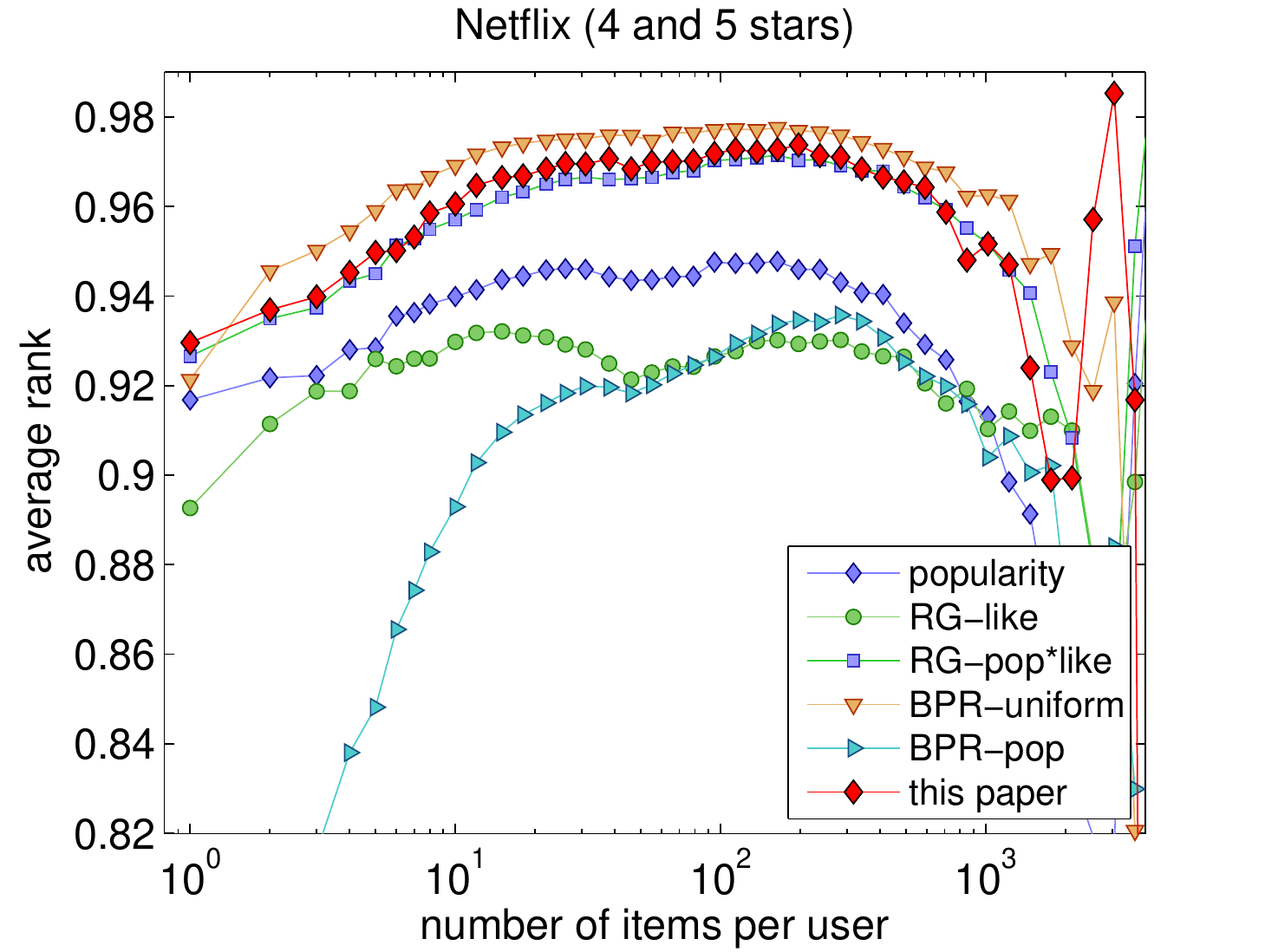}
\else
\emph{figure here}
\fi
\end{center}
\caption{The rank $R(i,j^{\star})$ in (\ref{eq:rank}), averaged over users and grouped logarithmically by $c_i$.
The \emph{top} evaluation is on the \emph{Xbox movies} sample, while the \emph{bottom} one is on the ``implicit feedback'' \emph{Netflix (4 and 5 stars)} set.
}
\label{fig:averagerank}
\end{figure}

A key application for modelling paired $(\textrm{\emph{user}}, \textrm{\emph{item}})$ symbols is large-scale recommendation systems, and we evaluate the predictions obtained by (\ref{eq:prediction}) on two large data sets.\footnote{Additional results follow in the Appendix \ref{sec:furthereval}.}
The \emph{Xbox movies} data is a sample of $5.6 \times 10^7$ views for $6.2 \times 10^6$ users on a sub-catalogue of around $1.2 \times 10^4$ movies \cite{Paquet_2013}. To evaluate on data known in the Machine Learning community, the four- and five-starred ratings from the Netflix prize data set were used to simulate a stream of ``implicit feedback'' $(\textrm{\emph{user}}, \textrm{\emph{item}})$ pairs in the \emph{Netflix (4 and 5 stars)} data.
We refer the reader to \cite{Paquet_2013} for a complete data set description.
For each user, one item was randomly removed to create a test set. To mimic a real scenario in the simplest possible way, each user's non-viewed items were ranked, and the position of the test item noted. We are interested in the rank of held out item $j^{\star}$ for user $i$ on a $[0,1]$ scale,
\begin{equation} \label{eq:rank}
R(i,j^{\star}) \defined \frac{1}{J - |\Gcal(i)|} \sum_{j \notin \Gcal(i)} \Ibb \Big[ f_{ij^{\star}} > f_{ij} \Big] \ ,
\end{equation}
where $f_{ij}$ indicates the score given by (\ref{eq:prediction}) or any alternative algorithm.

In Figure \ref{fig:averagerank}, we facet the average rank by $c_i$, the number of movie views per user.
As the evaluation is over 6 million users, this gives a more representative perspective than reporting a single average.
Apart from ranking by \emph{popularity} $c_j$, which would be akin to only factorizing with $s_i t_j$, we compare against two other baselines.
\emph{BPR-uniform} and \emph{BPR-pop} represent different versions of the Bayesian Personalized Ranking algorithm  \cite{Rendle_2009}, which optimizes a rank metric directly against either the data distribution of items (\emph{BPR-uniform}, with missing items are sampled uniformly during stochastic optimization), or a tilted distribution aimed at personalizing recommendations regardless of an item's popularity (\emph{BPR-pop}, with missing items sampled proportional to their popularity).
Their hyperparameters were set using cross-validation.
For the Random Graph model \cite{Paquet_2013}, rankings are shown with pure personalization (\emph{RG-like}) and with an item popularity rate factored in (\emph{RG-pop*like}).
The comparison in Figure \ref{fig:averagerank} is drawn using $K=20$ dimensions, $D' = D$ and hyperparameters set to one.
For \emph{Xbox movies}, the model outperforms all alternatives that we compared against.
\emph{BPR-uniform}, optimizing (\ref{eq:rank}) directly, performs slightly better on the less sparse Netflix set
(the Xbox usage sample is much sparser, as it is easier to rate many movies than to view as many).
For \emph{Xbox movies}, updating all item-related parameters in Algorithm \ref{alg} took 69 seconds on a 24-core (Intel Xeon 2.93Ghz) machine, and updating all user-related parameters took 83 seconds.

\begin{figure} [t!]
\begin{center}
\ifnum\figures=1
\includegraphics[width=0.48\textwidth]{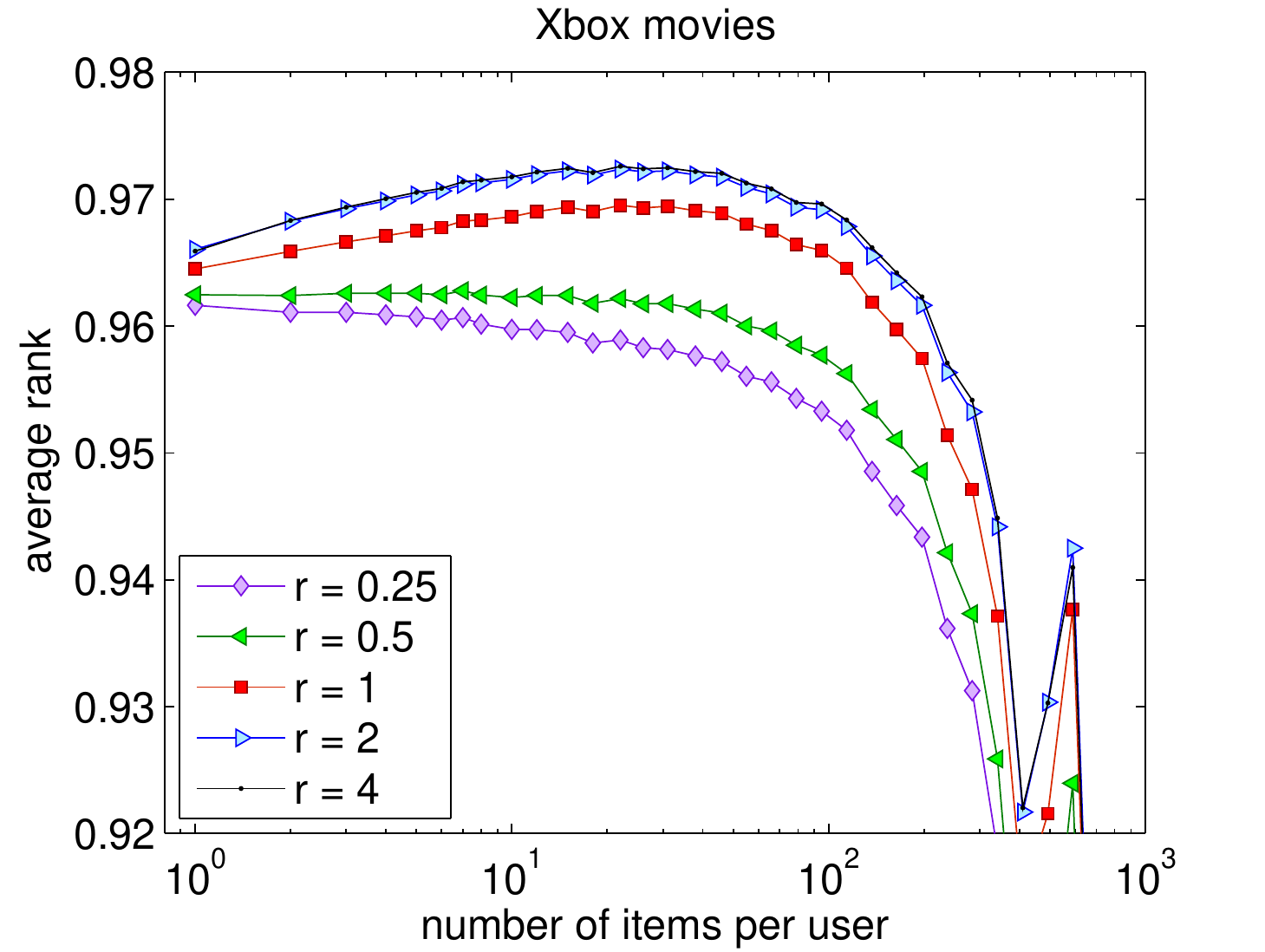}
\includegraphics[width=0.48\textwidth]{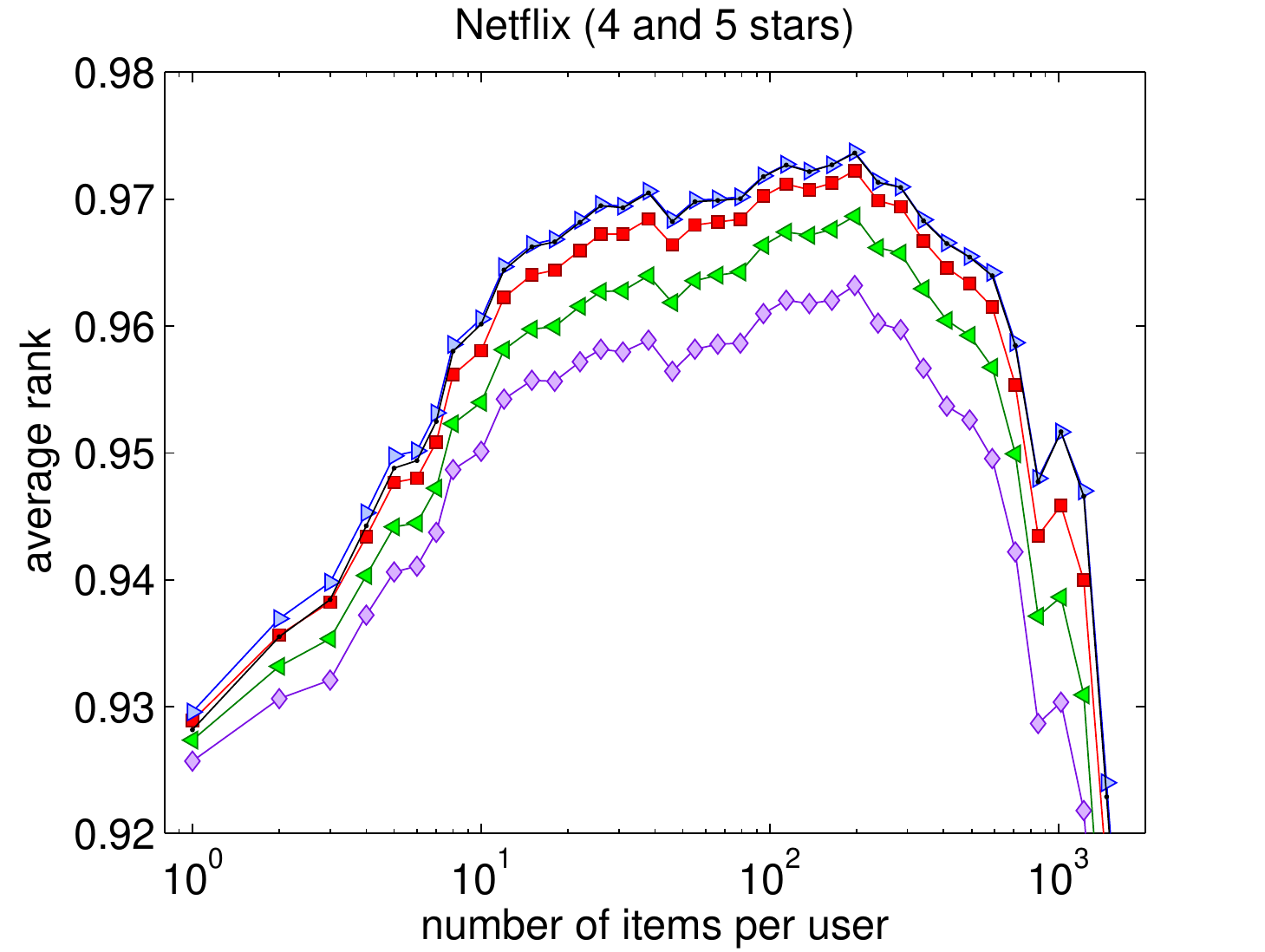}
\else
\emph{figure here}
\fi
\end{center}
\caption{The average rank $R(i,j^{\star})$ in (\ref{eq:rank}), grouped logarithmically by $c_i$, for varying values of $r$ in $D' = r D$.
}
\label{fig:D-values}
\end{figure}

\begin{figure*} [t]
\begin{center}
\ifnum\figures=1
\includegraphics[width=0.32\textwidth]{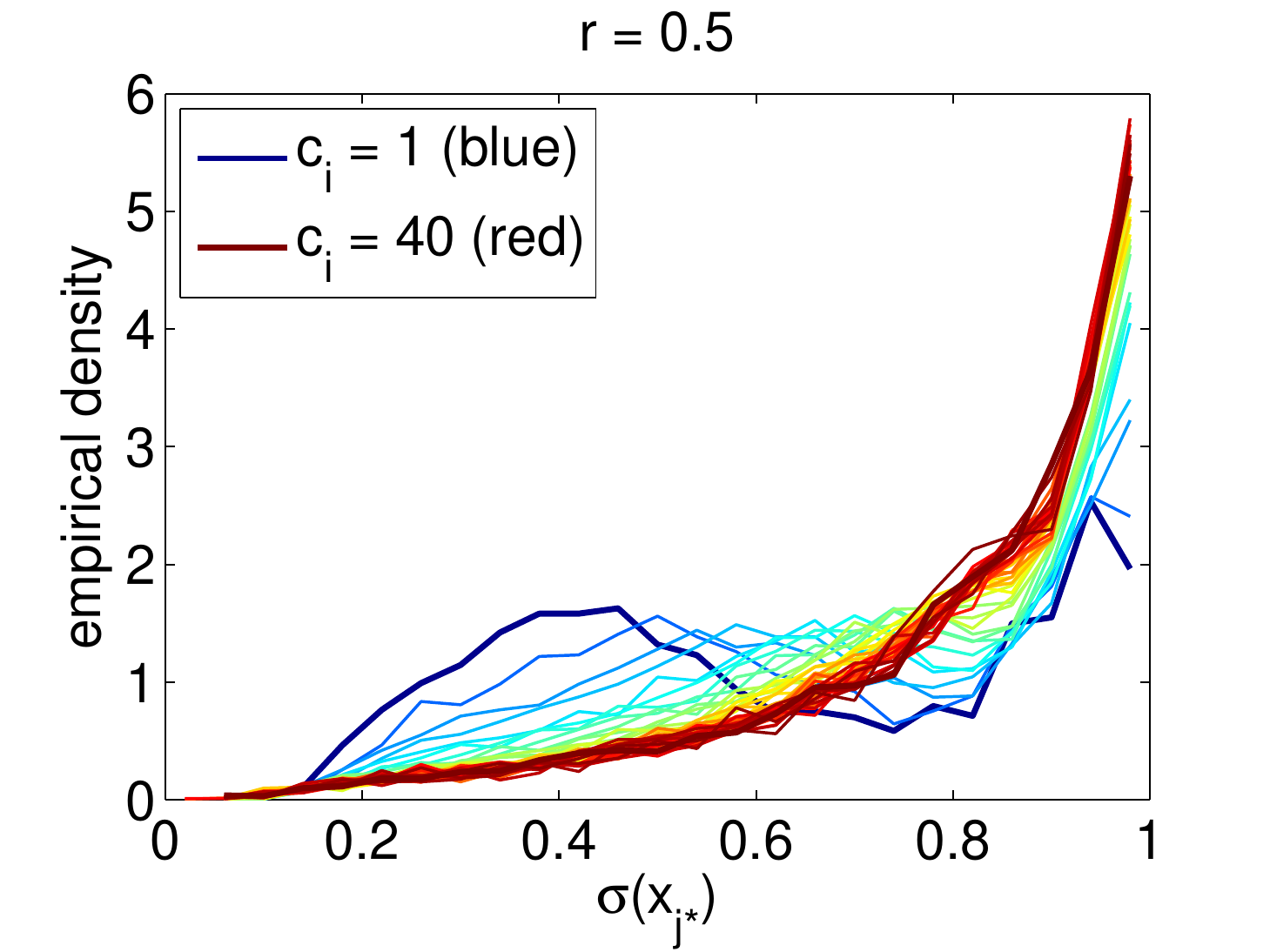}
\includegraphics[width=0.32\textwidth]{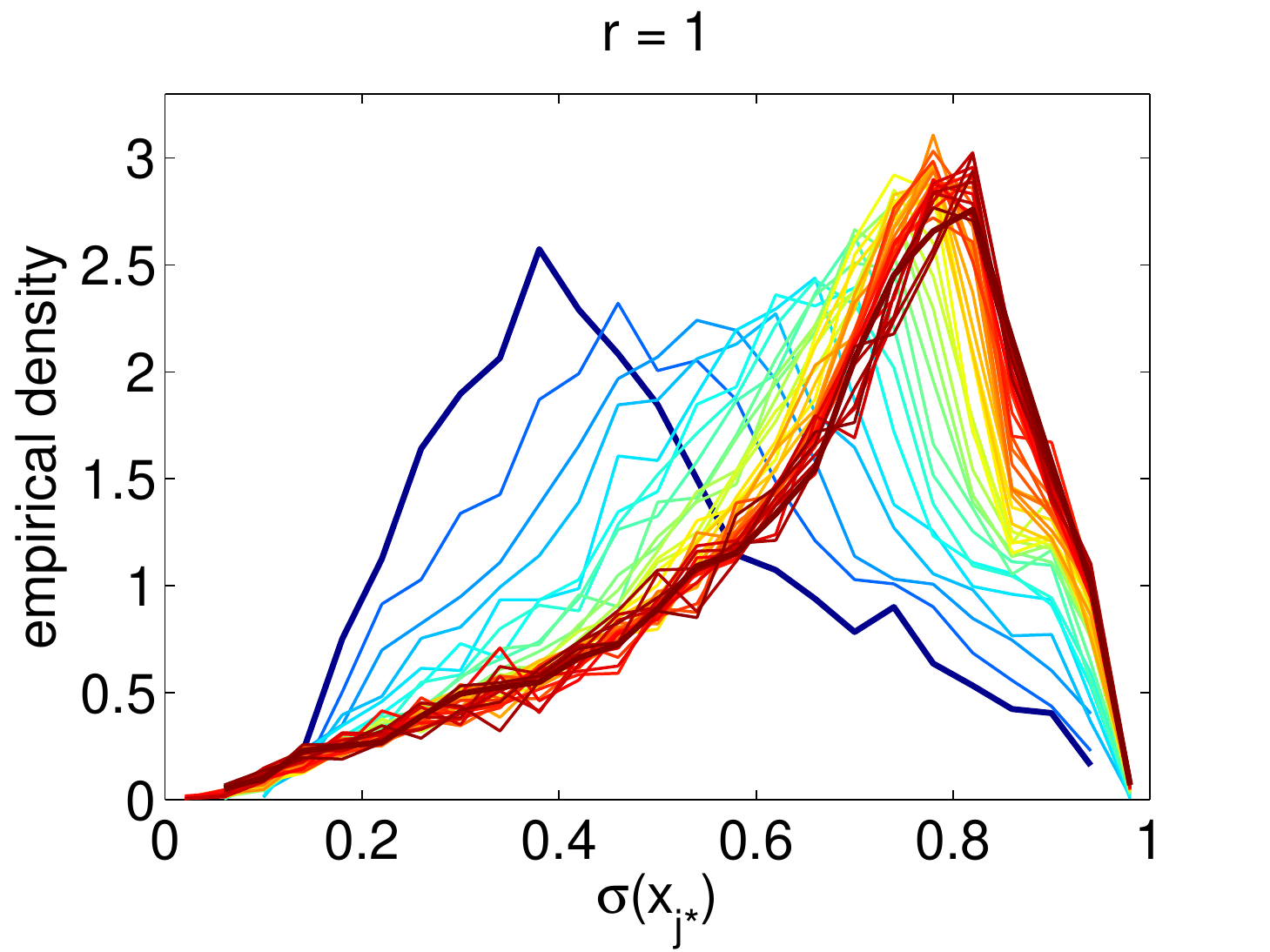}
\includegraphics[width=0.32\textwidth]{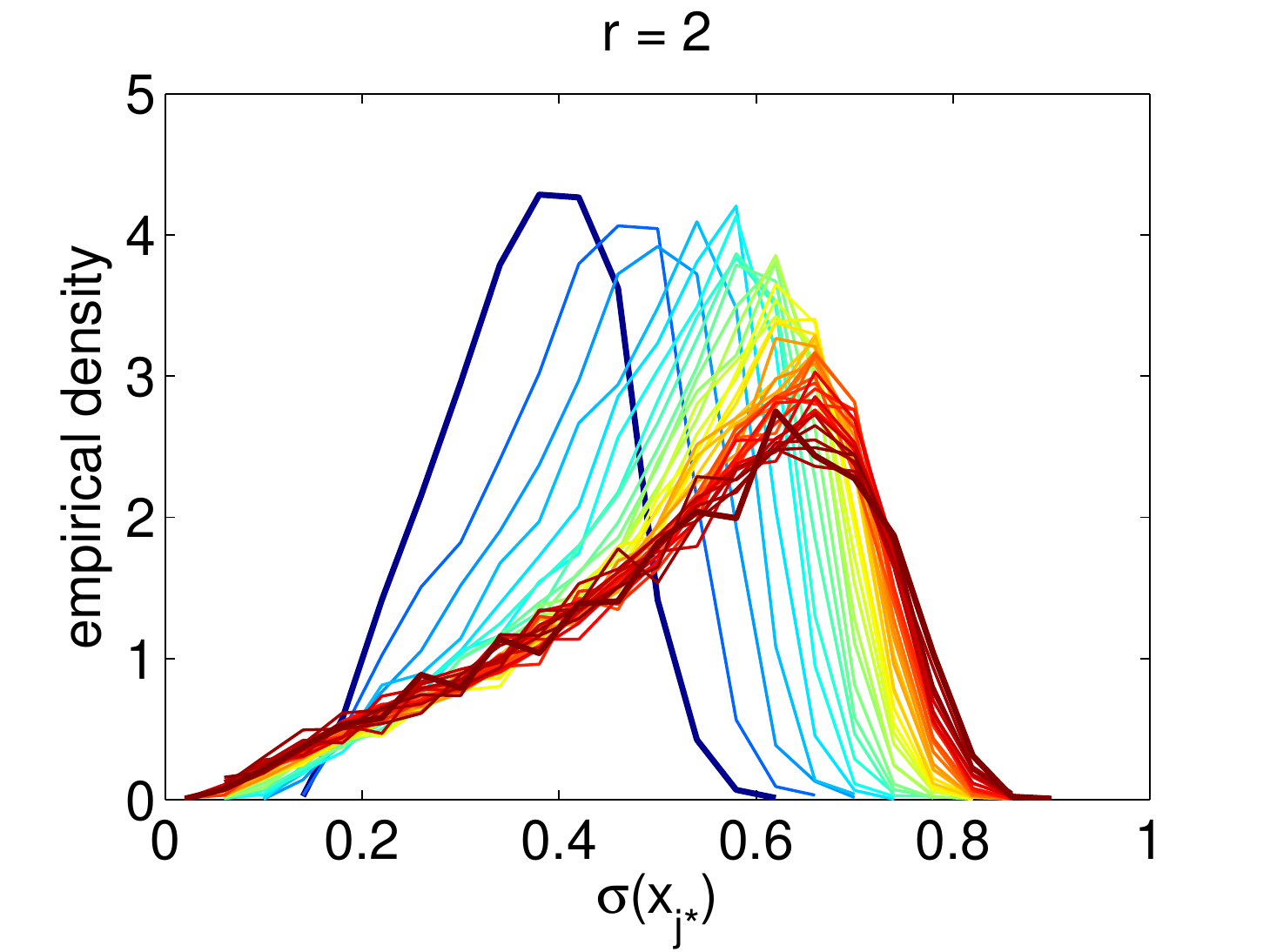}
\else
\emph{figure here}
\fi
\end{center}
\caption{The empirical densities of $\sigma(x_{j^{\star}})$, as defined in (\ref{eq:prediction}),
over all held-out items $j^{\star}$ in the \emph{Netflix (4 and 5 stars)} set. The densities are sliced according to $c_i = 1, \ldots, 40$ for different values of $r$ in $D' = r D$.
}
\label{fig:sigma}
\end{figure*}

We surmised in Section \ref{sec:scalable} that $D' = D$ is a reasonable hyperparameter setting, and Figure \ref{fig:D-values} validates this claim.
The figure shows the average held-out rank on the \emph{Netflix (4 and 5)} set for various settings of $D'$ through $D' =r D$.
The average rank improves beyond $r=1$, but empirically slowly decreases beyond $r=2$.
To provide insight into the \emph{``censoring''} step, Figure \ref{fig:sigma} accompanies Figure \ref{fig:D-values}, and shows the empirical density of the Bernoulli variable $\sigma(x_{j^{\star}})$ for held-out items $j^{\star}$. We break the empirical density down over users that appear in $c_i = 1, 2, 3, \ldots, 40$ pairs.
Given that the held-out pairs were observed, the Bernoulli variable should be \emph{true}, and the density of $\sigma(x_{j^{\star}})$ shifts right as $c_i$ becomes bigger.
The effect of having to explain less ($r = \frac{1}{2}$) or more ($r = 2$)
censored pairs is also visible in the figure.
There is also a slight benefit in increasing $K$. The average rank $\hat{R}_{20}$ for $K=20$ is 0.9649, using $r=1$. An increased latent dimensionality gives  $\hat{R}_{30} - \hat{R}_{20} = 1.07 \times 10^{-4}$, $\hat{R}_{40} - \hat{R}_{20} = 1.73 \times 10^{-4}$, and $\hat{R}_{50} - \hat{R}_{20} = 0.87 \times 10^{-4}$.

\section{Summary and outlook}

In this paper we presented a novel model for pairs of symbols, and showed state of the art results on a large-scale movies recommendation task.
Scalability was achieved by factorizing the popularity or selection step via $\pi_i \psi_j$, and employing ``site-independent'' variational bounds through careful parameter tying.
This approach might be too simplistic; an extension would be to use a $N$-component mixture model to select pairs with odds $\sum_{n=1}^{N} \pi_{in} \psi_{jn}$, and perform inference with Gibbs sampling.  

It is worth noting that B\"{o}hning \cite{Bohning_1992} and Bouchard \cite{Bouchard_2007} provide lower bounds to the logarithm of (\ref{eq:softmax}).
We originally embarked on a variational approximation to a posterior with (\ref{eq:softmax}) as likelihood using Bouchard's bound,
for which bookkeeping like Section \ref{sec:scalable}'s was done. However, with realistically large $I$ and $J$, solutions were trivial, as the means of the variational posterior approximations for $\u_i$ and $\v_j$ were zero. We leave B\"{o}hning's bound to future work.

\appendix

\section{The Joint Model}

The joint density in (\ref{eq:joint}) follows from combining the data likelihood
\begin{align*}
p(\data | \btheta)
& = \prod_d p(o_d = \true | \y_d, \z_d, \bvartheta) \, p(\y_d | \bpi) \, p(\z_d | \bpsi) \cdot
\prod_{d'} p(o_{d'} = \false | \y_{d'}, \z_{d'}, \bvartheta) \\
& = \prod_d \prod_{i,j} \big[ \pi_i \, \psi_j \, \sigma( \u_i^T \v_j+ b_i + b_j ) \big]^{y_{di} z_{dj}}
\prod_{d'} \prod_{i,j} (1 - \sigma( \u_i^T \v_j+ b_i + b_j ) )^{y_{d'i} z_{d'j}}
\end{align*}
with a prior on the unobserved variables $\btheta$, and rewriting the expression using 
observation counts $c_{ij} \defined \sum_d y_{di} z_{dj}$ for each pair $(i,j)$,
and marginal counts 
$c_i \defined \sum_d y_{di} $ and $c_j \defined \sum_d z_{dj}$.
The joint density is shown in Figure \ref{fig:graphicalmodel}.

\section{The Variational Bound} \label{sec:bounddetails}

For the sake of later derivations, it is worthwhile to explicitly write $\Lcal_{\bxi}[q]$ as it appears in (\ref{eq:L}).
It is
\begin{align}
\Lcal_{\bxi}[q]
& = \sum_{i,j} c_{ij} \, \Ebb_q \Bigg[ \log \sigma(\xi_{ij}) -
  \lambda(\xi_{ij}) \Big( ( \u_i^T \v_j + b_i + b_j )^2 - \xi_{ij}^2 \Big)
  + \frac{1}{2} ( \u_i^T \v_j + b_i + b_j ) - \frac{\xi_{ij}}{2} \Bigg] \nonumber \\
& \quad + \sum_{i,j} \sum_{d'} \Ebb_q[y_{d' i} z_{d' j}] \, \Ebb_q \Bigg[ \log \sigma(\xi_{ij})
  - \lambda(\xi_{ij})\Big( ( \u_i^T \v_j + b_i + b_j )^2 - \xi_{ij}^2 \Big) \nonumber \\
& \qquad\qquad\qquad\qquad\qquad\qquad\qquad\qquad\qquad\qquad\qquad\qquad
  - \frac{1}{2} ( \u_i^T \v_j + b_i + b_j ) - \frac{\xi_{ij}}{2} \Bigg] \nonumber \\
& \quad + \sum_i \left( c_i + \sum_{d'} \Ebb[y_{d'i}] \right) \Ebb_q[\log \pi_i]
  + \sum_j \left( c_j + \sum_{d'} \Ebb[z_{d'j}] \right) \Ebb_q[\log \psi_j] \nonumber \\
& \quad + \sum_i \sum_k \Ebb_q[\log p(u_{ik})] + \sum_j \sum_k \Ebb_q[\log p(v_{jk})]
  + \sum_i \Ebb_q[\log p(b_i)] + \sum_j \Ebb_q[\log p(b_j)] \nonumber \\
& \quad + \Ebb_q [\log p(\bpi)] + \Ebb_q [\log p(\bpsi)] \nonumber \\
& \quad - \sum_i \sum_k \Ebb_q[\log p(u_{ik})] - \sum_j \sum_k \Ebb_q[\log p(v_{jk})]
  - \sum_i \Ebb_q[\log p(b_i)] - \sum_j \Ebb_q[\log p(b_j)] \nonumber \\
& \quad - \Ebb_q [\log q(\bpi)] - \Ebb_q [\log q(\bpsi)] - \sum_{d'} \Ebb_q[\log p(\y_{d'})] - \sum_{d'} \Ebb_q[\log p(\z_{d'})] \ . \label{eq:thefulljoint}
\end{align}
All expectations are taken under $q(\btheta)$ defined in (\ref{eq:factorizing}).

\subsection{Bookkeeping}

The scalable parameter updates in Section \ref{sec:scalable}
rely on a number of cached quantities, which we state here together for completeness:
\begin{align*}
\left.\begin{aligned}
\P_{\ominus} & \defined \sum_{j} t_j \, \Ebb_{q} [ \v_j \v_j^T ] &
\m_{\ominus}^{\dagger} & \defined \sum_{j} t_j \Ebb_{q} [ b_j ] \Ebb_{q} [ \v_j ] & 
\m_{\ominus}^{\ddagger} & \defined \sum_{j} t_j \Ebb_{q} [ \v_j ] \\
\nu_{\ominus} & \defined \sum_{j} t_j \Ebb_{q} \big[ b_j \big] &
\varkappa_{\ominus} & \defined \sum_j t_j \Ebb_{q}[ b_j^2 ]
\end{aligned}
\right\}
\text{\emph{item}-background}
\\
\left.\begin{aligned}
\P_{\oplus} & \defined \sum_{i} s_i \, \Ebb_{q} [ \u_i \u_i^T ] &
\m_{\oplus}^{\dagger} & \defined \sum_i s_i \Ebb_{q} [ b_i ] \Ebb_{q} [ \u_i ] & 
\m_{\oplus}^{\ddagger} & \defined \sum_i s_i \Ebb_{q} [ \u_i ] \\
\nu_{\oplus} & \defined \sum_i s_i \Ebb_{q} \big[ b_i \big] &
\varkappa_{\oplus} & \defined \sum_i s_i \Ebb_{q}[ b_i^2 ]
\end{aligned}
\right\}
\text{\emph{user}-background}
\end{align*}

\subsection{Latent trait vector updates} \label{sec:traitupdates}

We stated $q(\btheta)$ in terms of the factorized Gaussian $\prod_k q(u_{ik})$, and will solve for $\prod_k q(u_{ik})$
by first maximizing an
intermediate lower bound with respect to the full-covariance Gaussian $\tilde{q}(\u_i) \defined \Ncal(\u_i ; \bmu_i, \P_i^{-1})$. Once $\tilde{q}(\u_i)$ is found,
a lower bound to it is maximized to find $\prod_k q(u_{ik})$.

\subsubsection{Scalable updates}

Let $\lambda_{ij} \defined \lambda(\xi_{ij})$.
The variational bound in (\ref{eq:thefulljoint}), as a function of the full-covariance Gaussian $\tilde{q}(\u_i)$, is
\begin{align}
\Lcal\big[\tilde{q}(\u_i)\big]
& = - \frac{1}{2} \sum_{j} c_{ij} \Bigg( 2 \lambda_{ij} \tr \left( \Ebb_{\tilde{q}} \big[ \u_i \u_i^T \big] \Ebb_q \big[ \v_j \v_j^T \big] \right) 
- 2 \Ebb_{\tilde{q}}[ \u_i ]^T \left( \frac{1}{2} - 2 \lambda_{ij} \Ebb_q[b_i + b_j] \right) \Ebb_q[ \v_j ] \Bigg)
\nonumber \\
& \quad - \frac{1}{2} \sum_{j} \sum_{d'} \Ebb_q[y_{d' i}] \, \Ebb_q[ z_{d' j}] \Bigg( 2 \lambda_{ij} \tr \left( \Ebb_{\tilde{q}} \big[ \u_i \u_i^T \big] \Ebb_q \big[ \v_j \v_j^T \big] \right)
\nonumber \\
& \qquad\qquad
- 2 \Ebb_{\tilde{q}}[ \u_i ]^T \left( - \frac{1}{2} - 2 \lambda_{ij} \Ebb_q[b_i + b_j] \right) \Ebb_q[ \v_j ] \Bigg) - \frac{1}{2} \tr \left( \Ebb_q \big[ \u_i \u_i^T \big] \tau_u \I \right) 
- \Ebb_{\tilde{q}} [\log \tilde{q}(\u_i)]
\nonumber \\
& = 
- \frac{1}{2}
\tr \Ebb_{\tilde{q}} \big[ \u_i \u_i^T \big] \left(\tau_u \I + 
\sum_{j} 2 \lambda_{ij} \left( c_{ij} +  \sum_{d'} \Ebb_q[y_{d' i}] \, \Ebb_q[ z_{d' j}] \right) \Ebb_q \big[ \v_j \v_j^T \big]
\right)
\nonumber \\
& \quad + \Ebb_{\tilde{q}}[ \u_i ]^T 
\sum_{j} \left( c_{ij} \left( \frac{1}{2} - 2 \lambda_{ij} \Ebb_q[b_i + b_j] \right) \phantom{\sum_j} \right.
\nonumber \\
& \qquad\qquad + \left.
\sum_{d'} \Ebb_q[y_{d' i}] \, \Ebb_q[ z_{d' j}] \left( - \frac{1}{2} - 2 \lambda_{ij} \Ebb_q[b_i + b_j] \right) \right)
\Ebb_q[ \v_j ] - \Ebb_{\tilde{q}} [\log \tilde{q}(\u_i)] \ ,
\label{eq:intermediatebound}
\end{align}
where $\tr$ denotes the $\trace$ operator.
$\Lcal[\tilde{q}(\u_i)]$ is maximized when $\tilde{q}(\u_i)$ is a Gaussian density $\Ncal(\u_i ; \bmu_i, \P_i^{-1})$
whose natural parameters $\P_i$ and $\m_i \defined \P_i \bmu_i$ are given by
(\ref{eq:Pi}) and (\ref{eq:mp}); they accompany 
$\Ebb_q \big[ \u_i \u_i^T \big]$ and $\Ebb_q[ \u_i ]$ in the quadratic and linear terms above.

The above expression contains a sum over $j = 1, \ldots, J$ and a further inner sum over $d' = 1, \ldots, D'$.
The scalable evaluation for $\P_i$ and $\m_i$ in Section \ref{sec:scalable}
incorporates caches $\P_{\ominus}$, $\m_{\oplus}^{\dagger}$, and $\m_{\oplus}^{\ddagger}$,
and only requires a sparse sum over $j \in \Gcal(i)$. The simplification is obtained by using
\begin{enumerate}
\item $\xi_{ij} = \xi^{*}$ (and hence $\lambda_{ij} = \lambda^{*}$) for all $j \notin \Gcal(i)$;
\item $c_{ij} = 0$ for all $j \notin \Gcal(i)$;
\item $\Ebb_q[y_{d' i}] = s_i$ for all $d' = 1, \ldots, D'$;
\item $\Ebb_q[z_{d' j}] = t_j$ for all $d' = 1, \ldots, D'$.
\end{enumerate}

\subsubsection{Intermediate bounds}

The bound $\Lcal[\tilde{q}(\u_i)]$ is maximized at
$\tilde{q}(\u_i) = \Ncal(\u_i ; \bmu_i, \P_i^{-1})$.
With $q'(u_{ik}) \defined \Ncal(u_{ik} ; \mu_{ik}, P_{i,kk}^{-1})$ being
the minimizer of the Kullback-Leibler divergence
\[
\prod_k q'(u_{ik}) \defined \argmin_{\prod_k q(u_{ik})}
\mathsf{D}_{\mathrm{KL}}\left(\prod_k q(u_{ik}) \Big\| \tilde{q}(\u_i) \right) \ ,
\]
we now show that $\Lcal[\tilde{q}(\u_i)]$ serves as a temporary or \emph{intermediate} lower bound to $\log p(\data)$:
\begin{equation} \label{eq:intermediate}
\tilde{\Lcal}\big[\tilde{q}(\u_i)\big] \ge \Lcal \left[ \prod_k q'(u_{ik}) \right] \ .
\end{equation}
The bound in (\ref{eq:intermediate}) follows by substituting 
$\Ebb_{\tilde{q}} [ \u_i \u_i^T ] = \bmu_i \bmu_i^T + \P_i^{-1}$ in (\ref{eq:intermediatebound}):
\begin{align*}
\tilde{\Lcal}\big[\tilde{q}(\u_i)\big]
& = -\frac{1}{2} \tr \Ebb_{\tilde{q}} \big[ \u_i \u_i^T \big] \P_i + \Ebb_{\tilde{q}}[ \u_i ]^T \P_i \bmu_i
- \Ebb_{\tilde{q}}\big[\log \tilde{q}(\u_i)\big] \\
& = - \frac{K}{2} + \frac{1}{2} \bmu_i^T \P_i \bmu_i - \left( - \frac{K}{2} \log(2 \pi \erm) + \frac{1}{2} \log |\P_i| \right) \ .
\end{align*}
Let $\diag(\P_i)$ indicate the $K$-by-$K$ matrix that contains only the diagonal of $\P_i$.
As $\Ebb_{q'} [ \u_i \u_i^T ] = \bmu_i \bmu_i^T + \diag(\P_i)^{-1}$ and $\tr \diag(\P_i)^{-1} \P_i = K$, the second bound expands as
\begin{align*}
\Lcal \left[ \prod_k q'(u_{ik}) \right]
& = -\frac{1}{2} \tr \Ebb_{q'} \big[ \u_i \u_i^T \big] \P_i + \Ebb_{q'} [ \u_i ]^T \P_i \bmu_i
- \Ebb_{q'}\left[\sum_k \log q'(u_{ik}) \right] \\
& = - \frac{K}{2} + \frac{1}{2} \bmu_i^T \P_i \bmu_i - \left(- \frac{K}{2} \log(2 \pi \erm) + \frac{1}{2} \log \big| \diag(\P_i) \big| \right) \ .
\end{align*}
Finally, (\ref{eq:intermediate}) follows from the identity $|\P_i| \le \prod_{k} P_{i,kk} = \big| \diag(\P_i) \big|$ as $\P_i$ is positive definite.

\subsubsection{The advantage of an intermediate bound}

By first solving for $\tilde{q}(\u_i)$, the updates in (\ref{eq:Pi}) and (\ref{eq:mp})
require \emph{one} sum over $j \in \Gcal(i)$, and an $\Ocal(K^3)$ matrix inverse to obtain $\bmu_i \defined \P_i^{-1} \m_i$
and $\prod_k q(u_{ik})$.
On the other hand, one may solve for each $q(u_{ik})$ for $k = 1, \ldots, K$ in turn. Each of these $K$ updates require a 
sum over $j \in \Gcal(i)$, but does not require the matrix inverse.
There is therefore a computational trade-off between these two options.
The trade-off depends on $|\Gcal(i)|$ and $K$, and wasn't investigated further in the paper.

\subsection{Logistic bound parameter updates}

All the $\xi_{ij}$ parameters are tied to $\xi^{*}$ for $(i,j) \notin \Gcal$, and we write 
(\ref{eq:thefulljoint}) as a function of $\xi^{*}$ as
\begin{align*}
\Lcal(\xi^{*})
& = \log \sigma(\xi^{*}) \sum_{(i,j) \notin \Gcal} D' s_i t_j
  - \lambda(\xi^{*}) \sum_{(i,j) \notin \Gcal} D' s_i t_j \Ebb_q \Big[  ( \u_i^T \v_j + b_i + b_j )^2 \Big]
\\
& \quad \quad + \left( \lambda(\xi^{*}) \, {\xi^{*}}^2 - \frac{\xi_{ij}}{2} \right) \sum_{(i,j) \notin \Gcal} D' s_i t_j \ .
\end{align*}
(Notice that for $(i,j) \notin \Gcal$ we have $c_{ij} = 0$, and $c_{ij}$ does not explicitly occur in the above expression.)
Recalling that
$\lambda(\xi) \defined \frac{1}{2 \xi} [\sigma(\xi) - \frac{1}{2}]$ and that $\sigma(\xi) \defined (1 + \erm^{-\xi})^{-1}$,
the above derivative is
\[
\frac{ \partial \Lcal(\xi^{*})}{ \partial \xi^{*}} = - \lambda'(\xi^{*})
\sum_{(i,j) \notin \Gcal} D' s_i t_j \Ebb_q \Big[  ( \u_i^T \v_j + b_i + b_j )^2 - (\xi^{*})^2 \Big] \ .
\]
As the bound is symmetric around $\xi^{*} = 0$ and as $\lambda'(\xi^{*})$ is a monotonic function of $\xi^{*}$ for $\xi^{*} \ge 0$, the derivative is zero when
\begin{equation} \label{eq:xi}
(\xi^{*})^{2} = \frac{1}{ \sum_{(i,j) \notin \Gcal} s_i t_j } \sum_{(i,j) \notin \Gcal} s_i t_j \Ebb_q \Big[  ( \u_i^T \v_j + b_i + b_j )^2 \Big] \ .
\end{equation}
Unfortunately (\ref{eq:xi}) requires a sum over $(i,j) \notin \Gcal$. However, (\ref{eq:locallogistic}) states that
$\xi_{ij}^2 = \Ebb_q[ (\u_i^T \v_j + b_i + b_j)^2]$ can be computed and discarded for $(i,j) \in \Gcal$ if required, and
hence the required sum can be written in terms of cached quantities through
\[
\sum_{(i,j) \notin \Gcal} s_i t_j \Ebb_q \Big[  ( \u_i^T \v_j + b_i + b_j )^2 \Big] = 
\sum_{i,j} s_i t_j \Ebb_q \Big[  ( \u_i^T \v_j + b_i + b_j )^2 \Big]
- 
\sum_{(i,j) \in \Gcal} s_i t_j \xi_{ij}^2 \ ,
\]
and using
\[
\sum_{i,j} s_i t_j \Ebb_q \Big[  ( \u_i^T \v_j + b_i + b_j )^2 \Big]
= \tr \P_{\oplus} \P_{\ominus} + 2 \m_{\oplus}^{\ddagger T} \m_{\ominus}^{\dagger}
+ 2 \m_{\oplus}^{\dagger T} \m_{\ominus}^{\ddagger}
+ 2 \nu_{\oplus} \nu_{\ominus} + \varkappa_{\oplus} + \varkappa_{\ominus}
\ .
\]

\subsection{Categorical updates}

We want to find $q(\y_{d'}) = \prod_{i=1}^{I} s_i^{y_{d' i}}$ which is a categorical distribution
parameterized by $\s$.
Using the notation
\begin{align*}
\Omega_{ij} & \defined \log \sigma(\xi_{ij}) - \lambda(\xi_{ij}) \Big( \Ebb_q[ (\u_i^T \v_j + b_i + b_j)^2] - \xi_{ij}^2 \Big)
 - \frac{\xi_{ij}}{2} - \frac{1}{2} \Ebb_q[\u_i^T \v_j + b_i + b_j]
\end{align*}
from Section \ref{sec:scalable}, substitute
$\Ebb_q[y_{d' i}] = s_i$ into (\ref{eq:thefulljoint}) to obtain $\Lcal_{\bxi}[q]$ as a function of $\s$:
\[
\Lcal^{l}(\s) = D' \sum_i s_i \sum_j t_j \Omega_{ij} + D' \sum_i s_i \Ebb_q[\log \pi_i]
- D' \sum_i s_i \log s_i + l\left( \sum_i s_i - 1 \right) \ .
\]
The above function includes a Lagrange multiplier $l$ as $\sum_i s_i$ normalizes to one.
The gradient of $\Lcal^{l}(\s)$ with respect to $s_i$ is zero when 
\[
\log s_i = \Ebb_q[\log \pi_i] + \sum_j t_j \Omega_{ij} + \frac{l}{D'} - 1 \ ,
\]
while the Lagrange multiplier gives the normalizer so that
\begin{equation} \label{eq:multinomial}
s_i = \frac{\erm^{ \Ebb_q[\log \pi_i] + \sum_j t_j \Omega_{ij} } }{
\sum_{i' = 1}^{I} \erm^{ \Ebb_q[\log \pi_{i'}] + \sum_j t_j \Omega_{i'j} } } \ .
\end{equation}

\subsubsection{Using caches}

Evaluating $\sum_j t_j \Omega_{ij}$ in (\ref{eq:multinomial}) for every $i = 1, \ldots, I$ 
again leaves us with an undesirable $\Ocal(IJ)$ complexity. Here, too, we shall make heavy use of cached quantities to
simplify this computation. First note that
\[
\sum_{j \notin \Gcal(i)} t_i \Omega_{ij}
= \sum_{j \notin \Gcal(i)} t_i \left( \log \sigma(\xi^{*}) - \lambda^{*} \Big( \Ebb_q[ (\u_i^T \v_j + b_i + b_j)^2] - (\xi^{*})^2 \Big)
 - \frac{\xi^{*}}{2} - \frac{1}{2} \Ebb_q[\u_i^T \v_j + b_i + b_j] \right)
\]
where $\lambda^{*} \defined \lambda(\xi^{*})$, and that $\sum_{j \notin \Gcal(i)} t_i \Omega_{ij} = \sum_{j \notin \Gcal(i)} t_i \Omega_{ij}^{*}$. We therefore compute the full sum $\sum_j t_j \Omega_{ij}^{*}$ using caches, and then only loop over the sparse set $j \in \Gcal(i)$ to incorporate the difference. That is,
\begin{align*}
\sum_j t_j \Omega_{ij}^* & =
- \lambda^* \Big( \tr \Ebb_q[ \u_i \u_i ^T ] \P_{\ominus} + 2 \Ebb_q[ b_i \u_i^T ] \m_{\ominus}^{\ddagger} 
 + 2 \Ebb_q[ \u_i^T ] \m_{\ominus}^{\dagger}
 + \Ebb_q[b_i^2] + 2 \Ebb_q[b_i] \nu_{\ominus} + \varkappa_{\ominus} \Big) \\
& \qquad \qquad + \log \sigma(\xi^*)
 +\frac{ (\xi^*)^2}{2} \lambda^*   - \frac{\xi^*}{2} - \frac{1}{2} \left( \Ebb_q[\u_i^T] \m_{\ominus}^{\ddagger} + \Ebb_q[b_i] + \nu_{\ominus} \right)
\end{align*}
is computed using bookkeeping, and finally
\[
\sum_j t_j \Omega_{ij} = \sum_{j\in \Gcal(i)} t_j (\Omega_{ij}-\Omega_{ij}^*) + \sum_j t_j \Omega_{ij}^*
\]
then relies on a sparse sum.

\section{Practical considerations}

\begin{figure} [t!]
\begin{center}
\ifnum\figures=1
\includegraphics[width=0.48\textwidth]{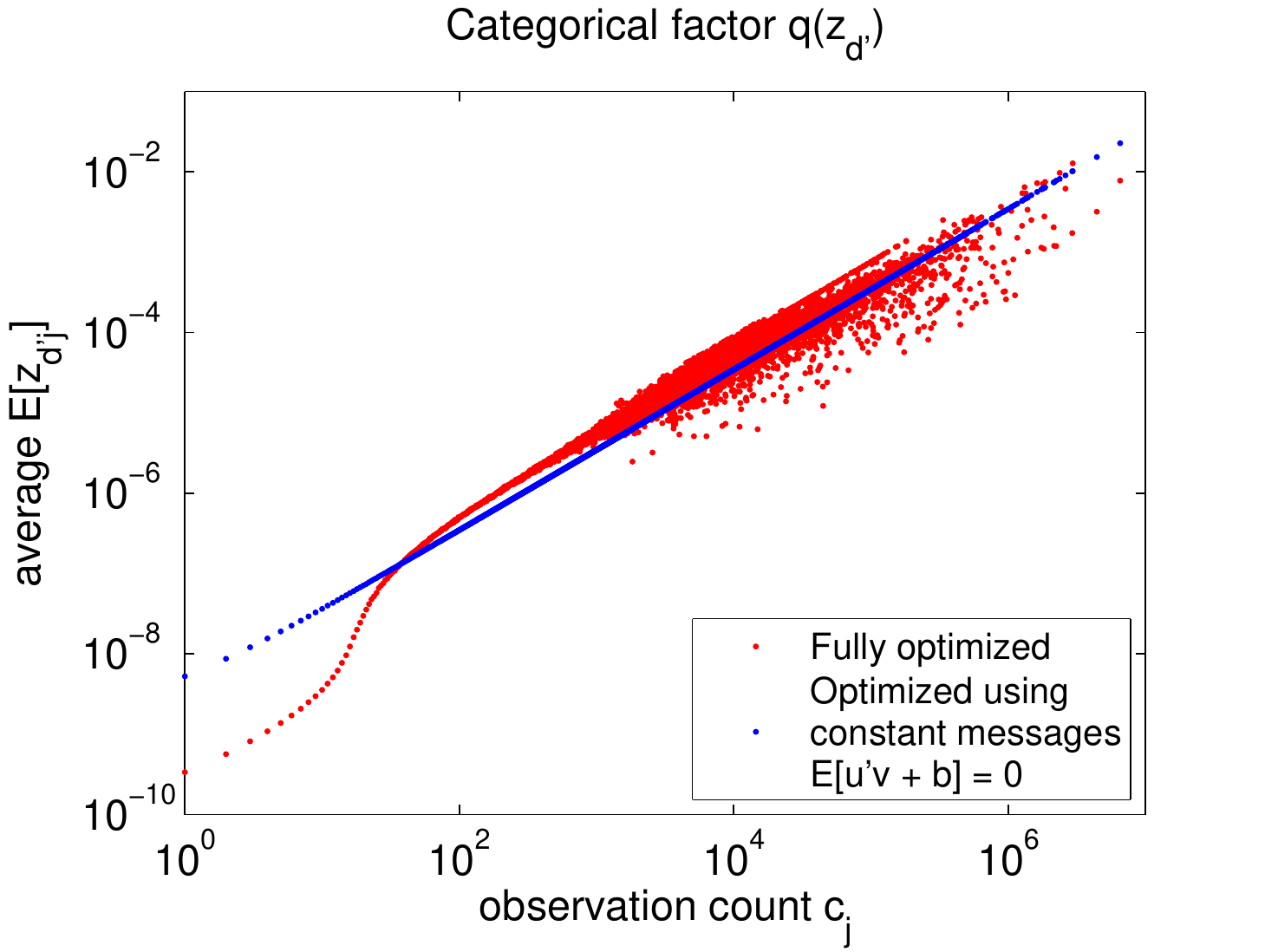}
\includegraphics[width=0.48\textwidth]{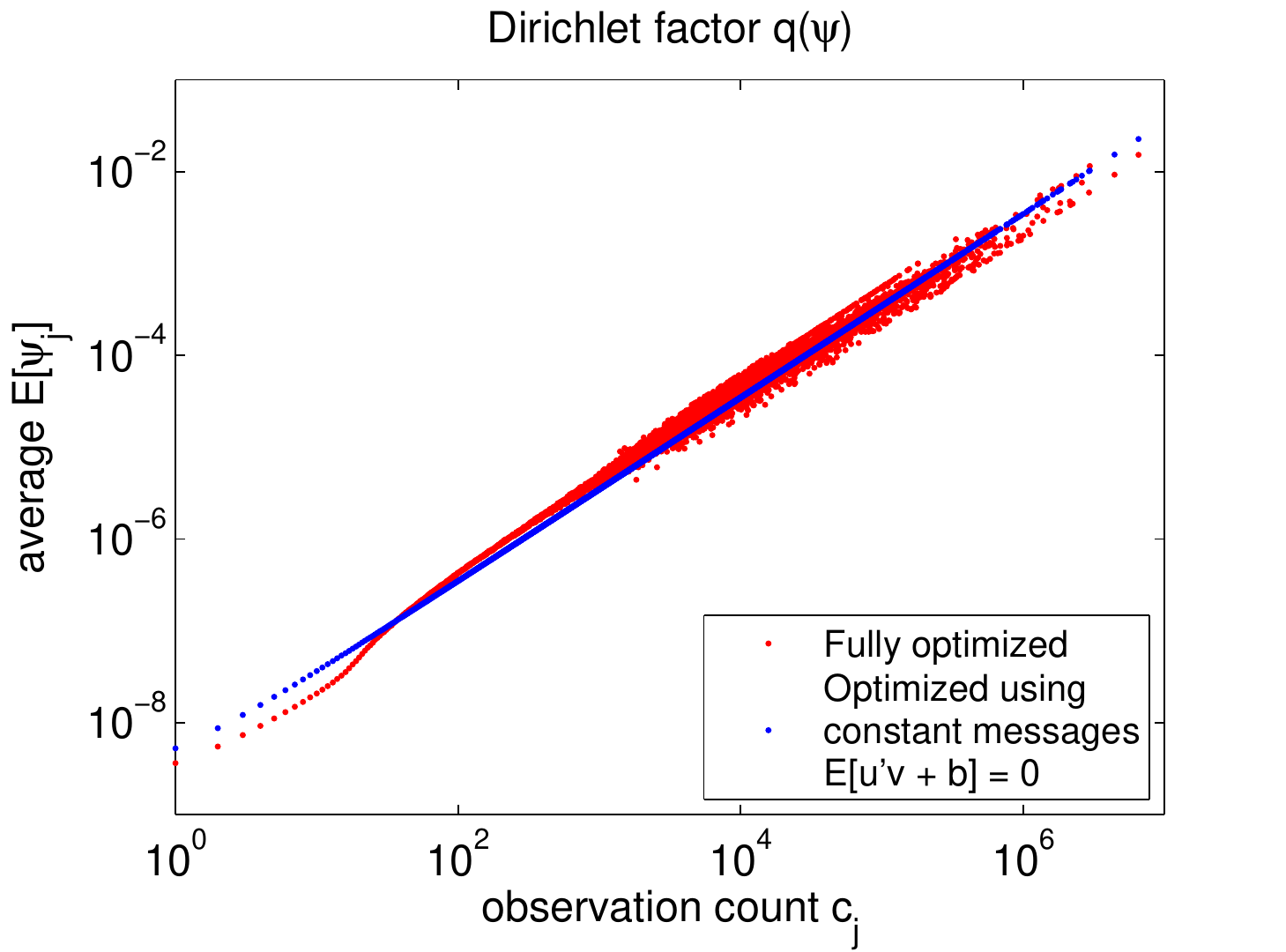}
\else
\emph{figure here}
\fi
\end{center}
\caption{The factors $q(\z_{d'})$ and $q(\bpsi)$ from a model with
$2.5 \times 10^8$ Windows 8 Phone App install signals as feedback pairs.
With ``variational model pruning'' of small components, a result that is more useful in a real system can be
created by using constant messages with $\Ebb[\u_i^T \v_j + b_i + b_j] = 0$ when optimizing for $q(\y_{d'})$ and $q(\z_{d'})$.
}
\label{fig:practical}
\end{figure}

``Variational model pruning'' \cite{MacKaySymmetryBreaking} can be observed on the fully optimized $q(\y_{d'})$ and $q(\z_{d'})$ factors.
In Figure \ref{fig:practical}, one sees the average $\Ebb[z_{d'j}]$ tailing roughly where $c_j < 50$. The net effect of disproportionately
decreasing the expected appearance probability is that $c_{ij}$ is explained by a much larger bias $b_j$.

In the context of the large-scale online system in which this model is deployed,
we've found it beneficial to substitute a constant $\Ebb[\u_i^T \v_j + b_i + b_j] = 0$ when optimizing for $q(\y_{d'})$ and $q(\z_{d'})$.

\section{Further evaluations} \label{sec:furthereval}

\begin{figure} [t]
\begin{center}
\ifnum\figures=1
\includegraphics[width=0.65\textwidth]{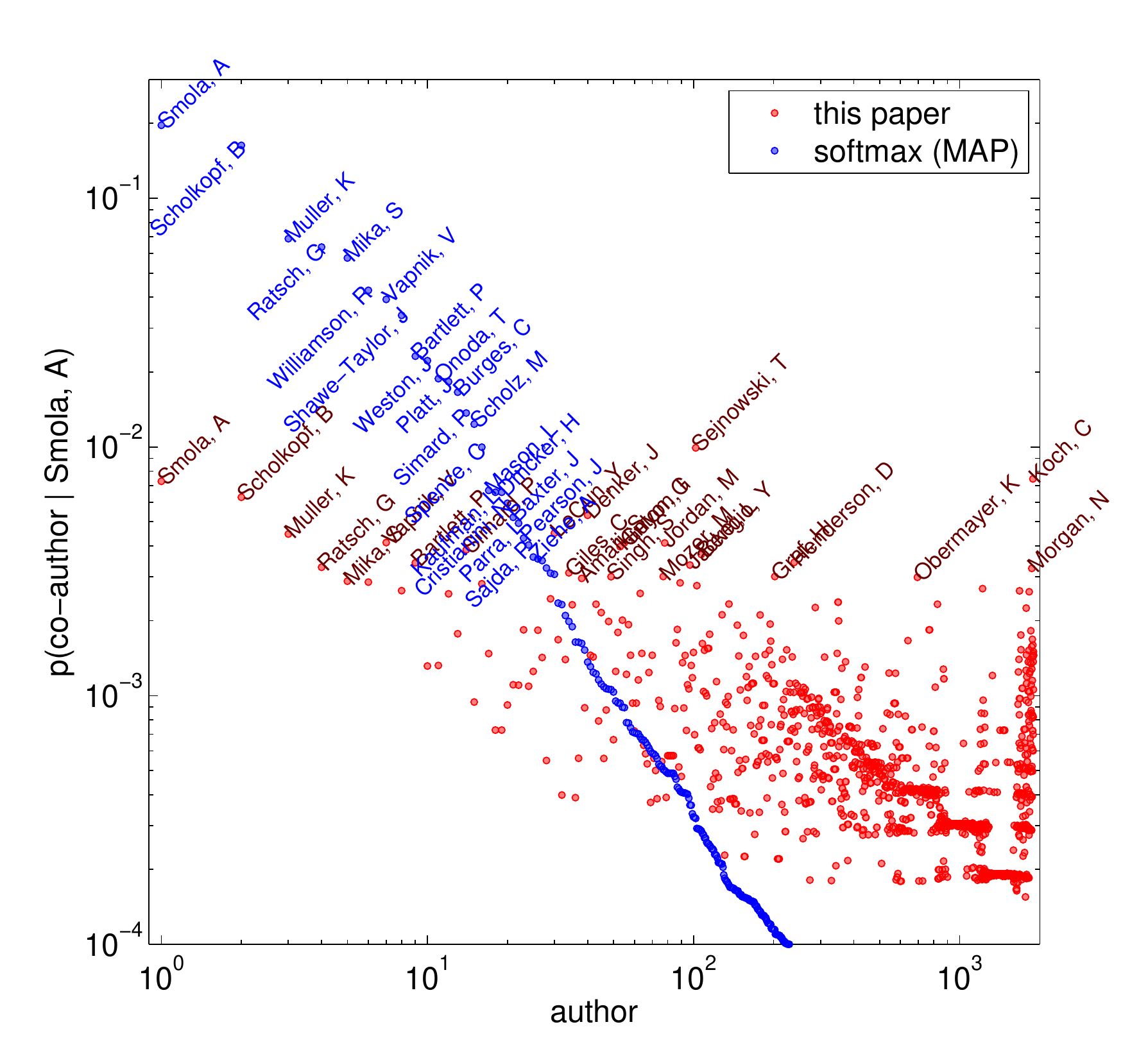}
\else
\emph{figure here}
\fi
\end{center}
\caption{A log-log plot comparing predictive densities obtained by this paper's model and a softmax equivalent.
Authors are ordered on the $x$-axis by the softmax MAP point estimate. The MAP estimate overfits with sparse data, as is evident in the (truncated blue) tail that approaches $10^{-8}$, and does not assign high odds to out-of-sample co-author K.~Obermayer, for example. 
}
\label{fig:Smola}
\end{figure}

As we do not directly maximize the softmax likelihood in (\ref{eq:softmax}), we are additionally interested in how the model's predictions differ from those obtained from a full softmax model. This is evaluated on a much smaller scale here.

\paragraph{Co-authorship networks} \label{sec:network}

This paper's starting point was the bilinear softmax likelihood function in (\ref{eq:softmax}). We will now turn to examine how much the approximation to $p(j | i, \data)$ in (\ref{eq:prediction}) deviates from that of a maximum a posteriori (MAP) solution to the softmax likelihood. 
As discussed earlier, the softmax MAP estimate is expensive to find, we thus use the relatively smaller NIPS 1--12 co-authorship dataset\footnote{\url{www.autonlab.org/autonweb/17433}} (even though it is not naturally bipartite data). 
We removed all single-authors which left us with $I = 1897$ authors, and treat co-authorship as \emph{symmetric} counts in $\data$.
Biases were included in (\ref{eq:softmax}), and excluded from our model, so that with $K=5$ both models have the same number of parameters.
We had to add the additional constraint $\u_i = \v_i$ for all $i$ to enforce the softmax point estimate to be symmetric.
This was not required for Algorithm \ref{alg}, which found a symmetric solution with and without such a constraint.
Figure \ref{fig:Smola} shows the predicted co-authors for A.~Smola,
with the top 25 predictions labelled for each model.
This is a density estimation problem with scarce data and an abundance of parameters, and with no shrinkage there are many singularities in the likelihood function.
With shrinkage ($\tau_u = 1$) the smallest softmax odds are $10^{-8}$ in Figure \ref{fig:Smola},
and the small data set is memorized by the MAP solution, which might not generalize.
This result underscores the need for a Bayesian approach.
Although the most probable predictions are still anecdotally interpretable,
we note that a truer comparison would be against posterior predictions that are estimated using Markov chain Monte Carlo samples with (\ref{eq:softmax}) as likelihood, but leave this research to future work.

\renewcommand\section{\subsubsection}

\small


\normalsize

\makeatletter
\renewcommand\section{\@startsection {section}{1}{\z@}{-2.0ex plus -0.5ex minus -.2ex}{1.5ex plus 0.3ex minus0.2ex}{\large\bf\raggedright}}
\makeatother

\end{document}